\documentclass[runningheads]{llncs}

\usepackage{accv}

\usepackage{accvabbrv}

\usepackage{graphicx}
\usepackage{booktabs}
\usepackage{microtype}
\usepackage[accsupp]{axessibility}  % Improves PDF readability for those with disabilities.

\usepackage{hyperref}

\usepackage{multirow}
\usepackage{pifont}
\newcommand{\cmark}{\ding{51}}  % ✓
\newcommand{\xmark}{\ding{55}}  % ✗

\newif\ifincludesupp
\includesupptrue

\begin{document}

\title{QATMA: Quantization-Aware Training with Multimodal Alignment for Open-Vocabulary Object Detection}
\titlerunning{QATMA: Quantization-Aware Training with Multimodal Alignment}

\author{Jinyeong Park\inst{1} \and Donghwa Kang\inst{2} \and Seunghwan An\inst{1} \and Insoo Kim\inst{1} \and Brent ByungHoon Kang\inst{2} \and Hyeongboo Baek$^\dagger$\inst{3} \and Jibum Kim$^\ddagger$\inst{1}}
\authorrunning{J. Park et al.}

\institute{Incheon National University, Incheon, South Korea \and
KAIST, Daejeon, South Korea \and
University of Seoul, Seoul, South Korea\\
Corresponding authors: $^\dagger$\email{hbbaek@uos.ac.kr}, $^\ddagger$\email{jibumkim@inu.ac.kr}}

\maketitle

\begin{abstract}
Quantizing open-vocabulary object detection (OVOD) models reduces their memory and computational costs, but extremely low-bit quantization severely degrades both cross-modal (region-text) and intra-modal (region-region) alignments. This multimodal degradation is a unique challenge that prior quantization methods for closed-vocabulary detectors fail to resolve. To overcome this, we propose Quantization-Aware Training with Multimodal Alignment (QATMA), the first multimodal-aware and architecture-agnostic QAT framework tailored for OVOD. QATMA integrates two key components: (i) Curriculum QAT, which partitions the detector by functional roles and progressively expands the quantization scope to suppress error accumulation and ensure stable optimization; and (ii) Text-anchored Pairwise Similarity Distillation, which transfers both region-text and region-region alignments from a full-precision teacher model via pairwise cosine similarities in the joint embedding space. Experimental results on LVIS and COCO zero-shot benchmarks demonstrate that QATMA significantly outperforms existing QAT baselines under extremely low-bit settings, achieving gains of up to 4.3 and 7.6 AP, respectively.
\end{abstract}
\section{Introduction}

The advent of vision-language models (VLMs)~\cite{CLIP,ALIGN} has shifted the object detection paradigm toward open-voca\-bu\-la\-ry object detection (OVOD)~\cite{OVRCNN}, which detects novel categories beyond predefined sets. However, OVOD models~\cite{GLIP,GLIPv2,ViLD,GroundingDINO} rely on heavy ViT-based backbones and text encoders, incurring massive computational overhead. Although lightweight, real-time OVOD models such as YOLO-World~\cite{YOLOWorld} have been proposed, their overhead remains substantial, so they still require further compression such as pruning~\cite{HanPruning,ChannelPruning}, knowledge distillation~\cite{KD,ChenDetKD}, or quantization~\cite{QAT,LSQ} for edge deployment.

\begin{figure}[tb]
    \centering
    \begin{minipage}[c]{0.60\linewidth}
        \begin{subfigure}[b]{\linewidth}
            \centering
            \includegraphics[width=\linewidth]{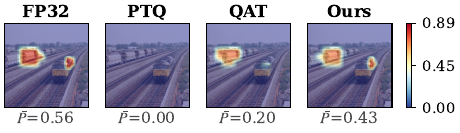}
            \caption{Region-text alignment: category ``Train''}
            \label{fig:obs_a}
        \end{subfigure}
        
        \begin{subfigure}[b]{\linewidth}
            \centering
            \includegraphics[width=\linewidth]{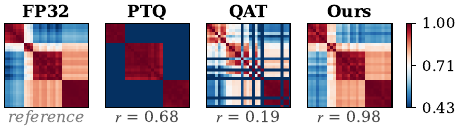}
            \caption{Region-region alignment}
            \label{fig:obs_b}
        \end{subfigure}
    \end{minipage}\hfill
    \begin{minipage}[c]{0.38\linewidth}
        \begin{subfigure}[b]{\linewidth}
            \centering
            \includegraphics[width=\linewidth]{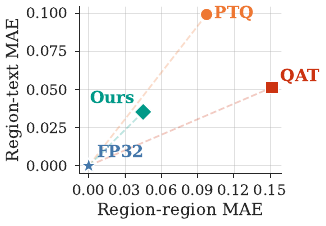}
            \caption{Embedding distortion}
            \label{fig:obs_c}
        \end{subfigure}
    \end{minipage}
    
    \caption{Impact of 4-bit quantization on YOLO-World~\cite{YOLOWorld} ( Objects365v2~\cite{Objects365}). (a) Confidence scores from the similarity between each region embedding and the ``Train'' text embedding ($\bar{P}$: mean over positive regions). (b) Pairwise cosine similarity matrix among positive region embeddings within the same category ($r$: Pearson correlation with FP32). (c) Quantitative comparison of embedding distortion relative to FP32; closer to the origin indicates less distortion.}
    \label{fig:observation}
\end{figure}

Quantization replaces floating-point operations with low-bit integer arithmetic without architectural changes. It is broadly categorized into post-training quantization (PTQ), which calibrates a pretrained model without retraining~\cite{DFQ,AdaRound,OMSE}, and quantization-aware training (QAT), which simulates quantization during training~\cite{QAT,LSQ,LSQPlus}.
While quantization has been explored for object detection~\cite{FQN,AQD,QDETR,OscQuant}, existing methods address only closed-vocabulary settings, leaving a critical gap for OVOD, which depends heavily on precise vision-language alignment.

We investigate this gap by analyzing extremely low-bit (4-bit) quantization on YOLO-World with the Objects365v2~\cite{Objects365} dataset. Specifically, we examine how quantization affects the fine-grained cross-modal and intra-modal alignment in the joint embedding space. The former is region-text alignment, which is the cosine similarity between region and text embeddings. The latter is region-region alignment, which is the pairwise cosine similarity among region embeddings assigned to the same text query. As shown in \cref{fig:observation}(a), PTQ destroys  region-text alignment, and naive QAT achieves only partial recovery. Furthermore, naive QAT fails to preserve region-region alignment (\cref{fig:observation}(b)).

A quantitative comparison (\cref{fig:observation}(c)) further reveals that naive QAT fails to jointly minimize the distortions of both region-text and region-region alignments, each measured as mean absolute error (MAE) from FP32. 
This suggests that, under extreme capacity constraints, the model overfits to ground-truth matching scores while sacrificing intra-modal alignment. Task loss alone is therefore insufficient to preserve both alignments. In contrast, our method jointly minimizes both distortions, most closely approaching the FP32 reference (\cref{fig:observation}(c)).
Knowledge distillation (KD) is a natural candidate for addressing this limitation, transferring knowledge from a full-precision teacher to a quantized student. However, at extremely low bit-widths, single-stage QAT is unstable~\cite{QDBEV}, and the large teacher--student capacity gap further hinders distillation~\cite{TAKD}.

In this context, we propose Quantization-Aware Training with Multimodal Alignment (QATMA), a novel framework that synergizes stage-by-stage optimization with module-specific knowledge distillation. QATMA is architecture-agnostic: it applies to any OVOD detector with the backbone-neck-head structure and region-text matching common to such models.
We design curriculum QAT (CQAT) to enable effective distillation by suppressing error accumulation. Instead of quantizing the whole detector at once, CQAT partitions it into functional modules and progressively expands the quantization scope, keeping the remaining modules frozen to isolate errors.

Building on this curriculum, we design a module-specific KD strategy tailored to each component's functional role. We apply feature distillation to the task-agnostic backbone to recover its representation capability. In contrast, the task-relevant neck-head governs the cross-modal and intra-modal alignment that conventional, closed-vocabulary detection KD does not address. We therefore propose Text-anchored Pairwise Similarity Distillation (TPSD), which jointly transfers region-text and region-region alignment via pairwise cosine similarities in the joint embedding space. 

Our main contributions are summarized as follows:
\begin{itemize}
    \item To the best of our knowledge, this is the first work to tackle extremely low-bit quantization of OVOD models. We propose QATMA, the first multimodal-aware and architecture-agnostic QAT framework, in which CQAT suppresses error accumulation by progressively quantizing the detector's functional modules, providing a stable optimization foundation that single-stage QAT lacks.
    \item We propose TPSD, a distillation method for the cross-modal and intra-modal alignment central to OVOD, which conventional detection KD overlooks. TPSD distills both alignments from the teacher through text-anchored pairwise cosine similarities in the joint embedding space.
    \item Extensive experiments demonstrate that QATMA significantly outperforms existing QAT baselines under extremely low-bit settings, with gains of up to 4.3 AP on LVIS and 7.6 AP on COCO. Beyond YOLO-World, QATMA also proves effective on the DETR-based OmDet-Turbo, confirming its generality across OVOD architectures.
\end{itemize}
\section{Related Work}

\subsection{Open-Vocabulary Object Detection}
Object detection has advanced through two-stage~\cite{RCNN,FasterRCNN}, one-stage~\cite{SSD,RetinaNet,FCOS}, transformer-based~\cite{DETR,DeformableDETR,DINODETR}, and YOLO-series~\cite{YOLO,YOLOv3,YOLOv7,YOLOv8} paradigms, but all rely on a fixed set of training categories~\cite{COCO,Objects365}. This closed-vocabulary assumption prevents novel category detection, restricting their open-world applicability.

To overcome this, OVOD was introduced to detect novel categories via arbitrary text queries. Early work like OVR-CNN~\cite{OVRCNN} established the standard OVOD setting. Leveraging large-scale VLMs~\cite{CLIP,ALIGN}, ViLD~\cite{ViLD}, RegionCLIP~\cite{RegionCLIP}, and OWL-ViT~\cite{OWLViT} transferred CLIP's visual-semantic representations, while GLIP~\cite{GLIP,GLIPv2} and Grounding DINO~\cite{GroundingDINO} reformulated detection as region-text matching. These methods achieve remarkable generalization, yet most rely on heavy ViT backbones (\eg, Swin-T~\cite{SwinTransformer}) and complex cross-modal fusion, incurring massive computational costs and memory footprints that hinder real-time deployment.

Recently, real-time OVOD detectors have emerged across diverse architectures, pre-computing text embeddings offline to remove the inference-time text encoder. For instance, YOLO-World~\cite{YOLOWorld} builds on a lightweight YOLOv8~\cite{YOLOv8} backbone, while OmDet-Turbo~\cite{OmDetTurbo} adopts DETR with an efficient fusion head. Despite this progress, such lightweight models still retain a substantial number of parameters and involve intensive cross-modal operations, necessitating further compression such as quantization for edge deployment. It remains a critical challenge to design an effective quantization technique that preserves the fine-grained vision-language alignment of OVOD models.

\subsection{Quantization for Object Detection}
Network quantization reduces memory and computational overhead by representing models in low-bit precision. Broadly categorized into PTQ~\cite{DFQ,AdaRound,OMSE} and QAT~\cite{QAT,DoReFaNet,PACT,LSQ,LSQPlus}, it has achieved notable success in image classification. While object detection quantization has been explored, it remains less investigated. Early methods~\cite{QAT,FQN,AuxQuant} were limited to 8-bit precision, struggled with training instability, or imposed architectural constraints. To achieve accurate, fully quantized object detection, subsequent works proposed multi-level batch normalization~\cite{AQD} and weight oscillation mitigation~\cite{OscQuant,OscFreeQuant}.

Since naive QAT struggles to recover performance in extremely low-bit settings, integrating knowledge distillation (KD) from a full-precision teacher has become prevalent~\cite{QKD,DistillQuant}. For dense prediction, task-specific designs are more effective, such as FPN-level feature distillation for one-stage detectors~\cite{QFD}, distribution rectification distillation for DETR-based detectors~\cite{QDETR}, and view-guided distillation for BEV-based 3D detection~\cite{QDBEV}.

However, all aforementioned studies are strictly confined to closed-vocabulary settings. The quantization of OVOD models, which heavily rely on fine-grained vision-language alignment, remains largely unexplored. To bridge this gap, our work proposes QATMA, a low-bit quantization framework tailored for OVOD models.

\section{Preliminaries}

In this section, we briefly review the fundamental concepts of network quantization and open-vocabulary object detection, which form the basis of our proposed framework.

\subsection{Quantization}\label{sec:quantization}

We assume uniform quantization for weights and activations. For a parameter $w \in \mathbb{R}$, $b$-bit uniform quantization maps it to a discrete level. Given a scale $s > 0$, a zero-point $z \in \mathbb{Z}$, and a range $[l, u]$ defined by the bit-width $b$, the process is defined as:
\begin{equation}
    \bar{w} = \left\lfloor \text{clip}\!\left(\frac{w}{s} + z,\; l,\; u\right) \right\rceil,
    \label{eq:quantize}
\end{equation}
where $\lfloor \cdot \rceil$ denotes round-to-nearest, 
and $\text{clip}(x, l, u)$ clips the value of $x$ to the interval $[l, u]$. 
The range is $[-2^{b-1}, 2^{b-1}-1]$ for signed and $[0, 2^{b}-1]$ for unsigned integers. The dequantized parameter $\hat{w}$ is reconstructed as:
\begin{equation}
    \hat{w} = (\bar{w} - z) \cdot s.
    \label{eq:dequantize}
\end{equation}
This scheme is termed symmetric quantization when $z = 0$, and asymmetric otherwise.

QAT simulates these operations during training. In the forward pass, $w$ undergoes quantize-dequantize steps (\cref{eq:quantize,eq:dequantize}) to compute the loss $\mathcal{L}$. During backpropagation, the straight-through estimator (STE)~\cite{STE} approximates gradients through the non-differentiable rounding operator. Furthermore, $s$ and $z$ can be jointly optimized as learnable parameters~\cite{LSQ,LSQPlus}, allowing the model to actively compensate for quantization errors, thereby achieving significant gains over post-training quantization in low-bit settings.

\subsection{Open-Vocabulary Object Detection}\label{sec:ovod}

Closed-vocabulary detectors rely on a parameterized classifier tailored to a fixed set of training categories $\mathcal{C}_{\text{base}}$. Consequently, they cannot detect novel categories in open-world scenarios. OVOD transcends this by detecting unseen categories $\mathcal{C}_{\text{novel}}$ ($\mathcal{C}_{\text{base}} \cap \mathcal{C}_{\text{novel}} = \emptyset$) during inference, despite training solely on $\mathcal{C}_{\text{base}}$ annotations~\cite{OVRCNN}.
 
A key idea behind OVOD is to replace the fixed classifier with text embeddings. Specifically, OVOD models~\cite{GLIP,GLIPv2,YOLOWorld} employ dual encoders, where a text encoder encodes each category query into a text embedding $\mathbf{t} \in \mathbb{R}^D$ and an image encoder extracts a region embedding $\mathbf{v} \in \mathbb{R}^D$ for each object candidate from multi-scale visual features.

Final classification relies on region-text similarity scores, computed as the cosine similarity between the region embedding $\mathbf{v}$ and the text embedding $\mathbf{t}$. Leveraging text encoders pretrained on massive datasets~\cite{CLIP}, this region-text matching seamlessly extends detection to both $\mathcal{C}_{\text{base}}$ and $\mathcal{C}_{\text{novel}}$ simply by altering the query set.

Given an input image $x$ and ground-truth annotations $y = \{(\beta_i, c_i)\}_{i=1}^{N_{\text{gt}}}$, where $N_{\text{gt}}$ is the number of ground-truth objects with bounding boxes $\beta_i$ and base categories $c_i \in \mathcal{C}_{\text{base}}$, the model optimizes:
\begin{equation}
    \mathcal{L}_{\text{task}}(x, y) = \mathcal{L}_{\text{cls}}(x, y) + \mathcal{L}_{\text{loc}}(x, y),
\end{equation}
where $\mathcal{L}_{\text{cls}}$ and $\mathcal{L}_{\text{loc}}$ denote the classification loss (\eg, focal loss~\cite{RetinaNet}) on region-text similarity scores and the localization loss (\eg, IoU-based losses~\cite{GIoU,FCOS}), respectively.

Crucially, the open-vocabulary detection capability relies heavily on the fine-grained alignment between region and text embeddings. 
Extremely low-bit quantization distorts this embedding space, degrading both cross-modal and intra-modal alignments and consequently impairing classification. Preserving these alignments under such quantization is the central motivation for QATMA.

\section{Method}

QATMA is an integrated framework that mitigates quantization error accumulation and restores both cross-modal and intra-modal alignments. It consists of two synergistic components: CQAT for stable, stage-by-stage optimization, and TPSD tailored for task-relevant modules to reconstruct region-text and region-region alignment.

\begin{figure}[tb]
    \centering
    \includegraphics[width=\textwidth]{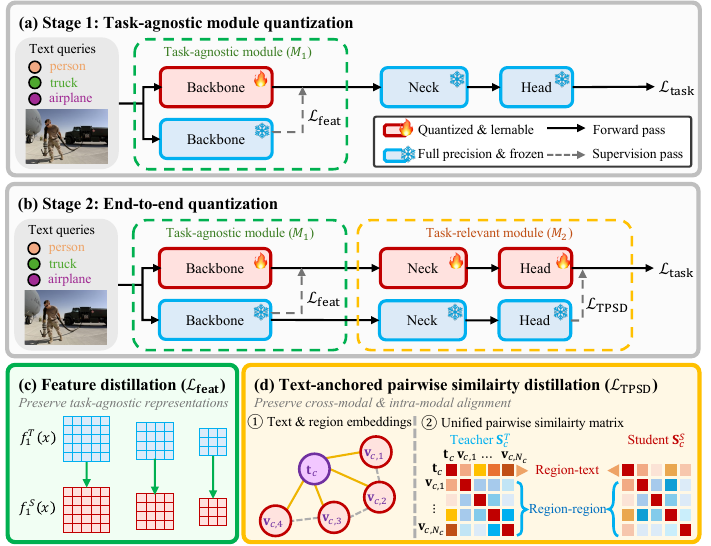}
    \caption{Overview of the proposed QATMA framework. (\emph{Red}) blocks denote quantized and learnable modules, and (\emph{blue}) blocks denote full-precision and frozen modules. (a) Stage 1: the backbone ($M_1$) is quantized with $\mathcal{L}_{\text{feat}}$ supervision from the full-precision teacher, while the neck-head remains frozen for error isolation. (b) Stage 2: the neck-head ($M_2$) is additionally quantized, supervised by both $\mathcal{L}_{\text{feat}}$ and $\mathcal{L}_{\text{TPSD}}$. (c) Feature distillation aligns the student's multi-scale backbone features $f_1^S(x)$ to those of the teacher $f_1^T(x)$. (d) TPSD groups region embeddings by text query $\mathbf{t}_c$ and constructs a unified pairwise similarity matrix $\mathbf{S}_c$ to transfer region-text and region-region alignment.}
    \label{fig:overview}
\end{figure}

\subsection{Curriculum QAT}\label{sec:cqat}

Single-stage QAT quantizes all layers simultaneously to optimize the task loss $\mathcal{L}_{\text{task}}$. However, in extremely low-bit settings, quantization errors originating from early layers accumulate rapidly and propagate through subsequent layers. This results in severe degradation of representation capability, making recovery challenging~\cite{QDBEV}.

To overcome this, CQAT partitions the model into $K$ functional modules $\{M_1,$ $M_2,$ $ \dots,$ $M_K\}$ and progressively expands the quantization scope along the data flow. Unlike prior progressive quantization, which proceeds at the level of individual weights~\cite{INQ} or layers~\cite{PROFIT}, CQAT instead defines the progression over functional modules grouped by their role in the detection pipeline, letting each stage be paired with a module-specific distillation objective (\cref{sec:trkd}).
Let $\mathbf{w}_k$ be the weights and $\phi_k$ be the quantization parameters (\eg, scaling factor, zero-point) of module $M_k$. At stage $k$, we apply QAT to the first $k$ modules while freezing the remaining modules ($M_{k+1}, \dots, M_K$) in full precision. 
The task loss at stage $k$ is formulated as:

\begin{equation}
\begin{aligned}
    \arg\min_{\{\mathbf{w}_i, \phi_i\}_{i=1}^{k}} &\quad \mathcal{L}_{\text{task}}(x,y) \\
    \text{s.t.} &\quad M_{k+1},\ldots,M_K \text{ remain in full precision}.
\end{aligned}
\end{equation}

The core principles underlying this design are \textit{error isolation} and \textit{sequential recovery}. 
Error isolation prevents premature noise propagation, allowing unquantized modules to maintain stable intermediate feature processing and gradient flow. Meanwhile, sequential recovery ensures that each module is optimized on top of quantization-adapted inputs from preceding modules, enabling it to primarily compensate for its own quantization error.

For widely-adopted OVOD architectures (\eg, GLIP~\cite{GLIP}, YOLO-World~\cite{YOLOWorld}), we instantiate a two-stage curriculum ($K=2$). As illustrated in \cref{fig:overview}, Stage 1 (\cref{fig:overview}(a)) quantizes the task-agnostic backbone ($M_1$) to recover from multi-scale feature degradation, utilizing the frozen neck-head ($M_2$) as an error isolator. Stage 2 (\cref{fig:overview}(b)) subsequently quantizes the neck-head to complete end-to-end optimization. We treat the neck and head as a single task-relevant module, as both carry out cross-modal fusion and region-text matching---the core multimodal operations in OVOD. We validate this grouping in \cref{sec:experiments}.

\subsection{Text-anchored Pairwise Similarity Distillation}\label{sec:trkd}

While CQAT structurally mitigates error accumulation, the information lost due to severe bit-width reduction remains substantial. To enhance sequential recovery, QATMA integrates module-specific KD into each curriculum stage. 
By leveraging the stable optimization foundation of CQAT, KD can be effectively applied even in extremely low-bit settings. Accordingly, we design a module-specific KD strategy based on the functional role of each module.

Given a full-precision teacher, the objective at stage $k$ is:
\begin{equation}
    \mathcal{L}_{\text{stage}\;k} = \mathcal{L}_{\text{task}} + \sum_{i=1}^{k} \lambda_i \mathcal{L}_{\text{KD}}^{(i)},
\end{equation}
where $\mathcal{L}_{\text{KD}}^{(i)}$ is the KD loss specifically designed for module $M_i$, and $\lambda_i$ is its balancing weight. As the curriculum progresses, KD losses from previous stages are retained as regularizers to prevent re-degradation.

\subsubsection{Backbone ($M_1$): Feature Mimicking.}
As a task-agnostic feature extractor, the backbone produces features whose quality fundamentally dictates the performance of all subsequent modules. Thus, we directly mimic the multi-scale features of the teacher's backbone (\cref{fig:overview}(c)) using a feature distillation loss $\mathcal{L}_{\text{feat}}$:
\begin{equation}
    \mathcal{L}_{\text{KD}}^{(1)} = \mathcal{L}_{\text{feat}}\!\left(f_1^S(x),\; f_1^T(x)\right),
\end{equation}
where $f_1^S(x)$ and $f_1^T(x)$ denote the backbone output features of the student and teacher, respectively. In practice, we adopt PKD~\cite{PKD} as $\mathcal{L}_{\text{feat}}$.

\subsubsection{Neck-head ($M_2$): TPSD.}
The task-relevant neck-head governs fine-grained vision-language alignment. Extremely low-bit quantization severely degrades both cross-modal and intra-modal alignments. Conventional KD, however, transfers intermediate features or output logits, rather than the alignment in the joint embedding space.
To address this, we propose TPSD, which uses text embeddings as anchors to capture both alignments within a unified pairwise similarity matrix (\cref{fig:overview}(d)).

For each text query $c$, TPSD constructs a matrix comprising its text embedding $\mathbf{t}_c$ and $N_c$ assigned region embeddings $\{\mathbf{v}_{c,n}\}_{n=1}^{N_c}$:
\begin{equation}
    \mathbf{X}_c = \begin{bmatrix} \mathbf{t}_c & \mathbf{v}_{c,1} & \cdots & \mathbf{v}_{c,N_c} \end{bmatrix}^\top \in \mathbb{R}^{(1+N_c) \times D},
\end{equation}
where $D$ is the embedding dimension. The region embeddings assigned to each text query (\ie, the positive regions for category $c$) are determined by the label assignment scheme of the base detector (\cref{sec:experiments}). 

After row-wise L2-normalization to obtain $\hat{\mathbf{X}}_c$, we compute the pairwise cosine similarity matrix:
\begin{equation}
    \mathbf{S}_c = \hat{\mathbf{X}}_c \hat{\mathbf{X}}_c^\top \in \mathbb{R}^{(1+N_c) \times (1+N_c)}.
\end{equation}
The first row and column of this matrix explicitly encode the region-text alignment, while the remaining internal blocks capture the region-region alignment. 

We compute this matrix for both the teacher and the student, denoted $\mathbf{S}_c^T$ and $\mathbf{S}_c^S$, respectively. By minimizing their discrepancy, the student retains the teacher's cross-modal and intra-modal alignment:
\begin{equation}
    \mathcal{L}_{\text{KD}}^{(2)} = \mathcal{L}_{\text{TPSD}} = \frac{1}{|\mathcal{C}|} \sum_{c \in \mathcal{C}} \frac{1}{(1+N_c)^2} \sum_{i=1}^{1+N_c} \sum_{j=1}^{1+N_c} \ell_\delta\!\left(S^S_{c,ij},\; S^T_{c,ij}\right),
\end{equation}
where $\mathcal{C}$ is the set of text queries in the image, and $\ell_\delta$ denotes the smooth $L_1$ loss. Since object frequencies are long-tailed, frequent or large-object categories yield far more assigned regions $N_c$, which would dominate a single global average. We therefore adopt a text-query-balanced average: normalizing within each text query before averaging across queries, so that every text query contributes equally regardless of $N_c$.

\section{Experiments}\label{sec:experiments}

QATMA is architecture-agnostic, making it directly applicable to any OVOD detector configured with a backbone-neck-head structure and region-text matching. We primarily evaluate our framework on YOLO-World~\cite{YOLOWorld}—a representative, lightweight, real-time OVOD detector—under an extremely low-bit (4-4-8) quantization setting for zero-shot detection, alongside comprehensive ablation studies. Additionally, we evaluate QATMA on OmDet-Turbo~\cite{OmDetTurbo} to verify its architectural generality.

\subsection{Experimental Settings}\label{sec:exp_settings}

\subsubsection{Datasets and Evaluation.}
For QAT training, we use the Objects365v2~\cite{Objects365} train split and GQA~\cite{GQA}.
Following YOLO-World, zero-shot evaluation is conducted on LVIS~\cite{LVIS} \texttt{minival} with the Fixed AP~\cite{FixedAP} and COCO~\cite{COCO} \texttt{val2017}.

\subsubsection{Quantization Settings.}
We use official pre-trained YOLO-World-M/L/X checkpoints. The text encoder (CLIP~\cite{CLIP}) is excluded from quantization, as it is removed at inference by pre-computing text embeddings offline~\cite{YOLOWorld}. We apply symmetric quantization for weights and asymmetric for activations and attention. Our main results adopt a 4-4-8 bit-width configuration (weight-activation-attention) with Ch-T-H granularity (per-channel weight, per-tensor activation, per-head attention). The first and last layers remain unquantized, following common practice~\cite{PACT,DoReFaNet}. For calibration, we use 256 samples from the Objects365v2 validation set.

\subsubsection{Training Details.}
We adopt the YOLO-World pretraining configuration, except for the learning rate of $1.5 \times 10^{-5}$ and the batch size of 48 for QAT. The learning rate scheduler is disabled as we fine-tune for 1 epoch. The text encoder remains frozen. Quantization parameters are learned via LSQ~\cite{LSQ,LSQPlus} with a learning rate of 0.1$\times$ the base rate. Curriculum stage 1 (backbone) uses the first 1/3 of the data, while stage 2 (neck-head) uses the remaining 2/3. The KD loss weight $\lambda_i$ is 6.0 for both feature distillation and TPSD. For TPSD, regions assigned to each text query are positive samples determined by the teacher's task-aligned label assignment (TAL)~\cite{TOOD}.

\begin{table}[tb]
    \caption{LVIS \texttt{minival} zero-shot evaluation with 4-4-8 quantization (Ch-T-H granularity). Ch = per-channel, T = per-tensor, H = per-head. Evaluation follows the LVIS fixed AP protocol~\mbox{\cite{FixedAP}}.}
    \label{tab:lvis-zeroshot}
    \centering
    \begin{tabular*}{\columnwidth}{@{\extracolsep{\fill}}llcclccccc@{}}
      \toprule
      Model & Bits & Size(MB) & BOPs(T) & Method & AP & AP$_r$ & AP$_c$ & AP$_f$ \\
      \midrule
      \multirow{5}{*}{YOLO-World-M}
        & FP32 & 110.88 & 51.82 & - & 31.0 & 23.8 & 29.2 & 33.9 \\
      \cmidrule(l){2-9}
        & \multirow{4}{*}{4-4-8} & \multirow{4}{*}{15.11} & \multirow{4}{*}{7.10} & QAT & 13.8 & 10.7 & 11.6 & 16.3 \\
        & & & & EMA+QC~\cite{OscQuant} & 13.7 & 11.8 & 12.2 & 15.4 \\
        & & & & QFD~\cite{QFD} & 13.0 & 7.1 & 11.0 & 15.8 \\
        & & & & Ours & \textbf{16.3} & 11.4 & \textbf{14.5} & \textbf{18.8} \\
      \midrule
      \multirow{5}{*}{YOLO-World-L}
        & FP32 & 181.4 & 98.39 & - & 35.4 & 27.6 & 34.1 & 38.0 \\
      \cmidrule(l){2-9}
        & \multirow{4}{*}{4-4-8} & \multirow{4}{*}{24.32} & \multirow{4}{*}{8.18} & QAT & 12.8 & 8.2 & 11.2 & 15.0 \\
        & & & & EMA+QC~\cite{OscQuant} & 13.1 & 9.0 & 11.3 & 15.3 \\
        & & & & QFD~\cite{QFD} & 13.0 & 9.5 & 11.9 & 14.7 \\
        & & & & Ours & \textbf{16.1} & \textbf{13.4} & \textbf{14.3} & \textbf{18.2} \\
      \midrule
      \multirow{5}{*}{YOLO-World-X}
        & FP32 & 281.11 & 149.08 & - & 36.6 & 29.4 & 35.0 & 39.4 \\
      \cmidrule(l){2-9}
        & \multirow{4}{*}{4-4-8} & \multirow{4}{*}{37.19} & \multirow{4}{*}{9.32} & QAT & 12.7 & 7.4 & 11.7 & 14.5 \\
        & & & & EMA+QC~\cite{OscQuant} & 12.0 & 11.7 & 9.4 & 14.4 \\
        & & & & QFD~\cite{QFD} & 13.2 & 9.7 & 11.6 & 15.3 \\
        & & & & Ours & \textbf{17.0} & \textbf{14.3} & \textbf{15.1} & \textbf{19.1} \\
      \bottomrule
    \end{tabular*}
\end{table}

\subsection{Main Results}\label{sec:main_results}

\begin{table}[tb]
    \caption{COCO \texttt{val2017} zero-shot evaluation with 4-4-8 quantization (Ch-T-H granularity)}
    \label{tab:coco-zeroshot}
    \centering
    \begin{tabular*}{\columnwidth}{@{\extracolsep{\fill}}llclcccccc@{}}
      \toprule
      Model & Bits & BOPs(T) & Method & AP & AP$_{50}$ & AP$_{75}$ & AP$_s$ & AP$_m$ & AP$_l$ \\
      \midrule
      \multirow{3}{*}{YOLO-World-M}
        & FP32 & 44.56 & - & 41.9 & 57.0 & 45.5 & 26.7 & 46.1 & 55.0 \\
      \cmidrule(l){2-10}
        & \multirow{2}{*}{4-4-8} & \multirow{2}{*}{2.05} & QAT & 22.0 & 32.1 & 23.6 & 7.8 & 23.5 & 32.4 \\
        & & & Ours & \textbf{26.1} & \textbf{38.1} & \textbf{28.1} & \textbf{11.0} & \textbf{28.1} & \textbf{37.9} \\
      \midrule
      \multirow{3}{*}{YOLO-World-L}
        & FP32 & 90.60 & - & 45.1 & 60.7 & 48.9 & 29.8 & 49.8 & 57.5 \\
      \cmidrule(l){2-10}
        & \multirow{2}{*}{4-4-8} & \multirow{2}{*}{3.09} & QAT & 18.4 & 27.2 & 19.6 & 7.5 & 20.1 & 26.9 \\
        & & & Ours & \textbf{25.2} & \textbf{36.7} & \textbf{27.0} & \textbf{10.5} & \textbf{27.5} & \textbf{37.7} \\
      \midrule
      \multirow{3}{*}{YOLO-World-X}
        & FP32 & 140.66 & - & 46.7 & 62.8 & 50.9 & 31.8 & 51.2 & 60.8 \\
      \cmidrule(l){2-10}
        & \multirow{2}{*}{4-4-8} & \multirow{2}{*}{4.21} & QAT & 18.6 & 26.8 & 19.9 & 7.8 & 20.5 & 27.0 \\
        & & & Ours & \textbf{26.2} & \textbf{38.2} & \textbf{28.1} & \textbf{12.2} & \textbf{28.7} & \textbf{37.2} \\
      \bottomrule
    \end{tabular*}
\end{table}

The 4-4-8 (W4A4) configuration with per-tensor activation quantization is highly aggressive.
PTQ suffers severe performance collapse (AP 0.0) across all models on both LVIS and COCO regardless of calibration strategy (MinMax, Percentile~\cite{FQN}, OMSE~\cite{OMSE}), demonstrating that low-bit quantization of OVOD models is infeasible without training.
The QAT baseline simultaneously quantizes all layers and trains with only the task loss $\mathcal{L}_{\text{task}}$, using LSQ~\cite{LSQ,LSQPlus}, a widely adopted learnable quantization scheme. For fair comparison, the baseline shares the same training configuration, total data, and number of iterations as QATMA.
We additionally compare against two closed-vocabulary detection QAT baselines: (i) EMA+QC~\cite{OscQuant}, which mitigates weight and activation oscillations during low-bit YOLO QAT, and (ii) QFD~\cite{QFD}, a KD-based QAT that quantizes a full-precision teacher's FPN-level features and distills them to the student.
Under this setting, 4-4-8 quantization reduces model size by up to 7.6$\times$ and bit operations (BOPs) by up to 33.4$\times$ (reported per benchmark in \cref{tab:lvis-zeroshot,tab:coco-zeroshot}, as BOPs vary with the number of text queries).

\subsubsection{LVIS MiniVal Zero-Shot.}
LVIS~\cite{LVIS} contains 1,203 object categories, each classified by annotation frequency into rare (AP$_r$), common (AP$_c$), and frequent (AP$_f$).
The large number of categories and the high proportion of rare categories make it a benchmark that directly reflects the difficulty of open-vocabulary detection.
\Cref{tab:lvis-zeroshot} reports results for YOLO-World-M/L/X.
While QAT recovers from the complete failure of PTQ, a large gap relative to FP32 remains, with drops of $-$17.2 AP on YOLO-World-M and $-$22.6 on YOLO-World-L.
QATMA consistently outperforms QAT across all models, with improvements of +2.5, +3.3, and +4.3 AP on YOLO-World-M, L, and X, respectively.
Notably, QAT degradation grows with model scale, and our improvements grow accordingly, indicating that our method becomes more beneficial as the capacity gap widens.
The gains are particularly pronounced in AP$_r$, which measures rare category detection and serves as a key indicator of open-vocabulary capability~\cite{GLIP,YOLOWorld}.
Since rare categories heavily depend on vision-language alignment quality, the substantial AP$_r$ gains of +5.2 on YOLO-World-L and +6.9 on YOLO-World-X, corresponding to 63.4\% and 93.2\% relative improvements, demonstrate that QATMA effectively restores the fine-grained alignment degraded by quantization.
Moreover, QATMA consistently outperforms not only naive QAT but also the closed-vocabulary detection QAT baselines EMA+QC and QFD across all model scales. Since these baselines do not target the vision-language alignment that underlies OVOD zero-shot detection, they fail to recover the alignment distorted under 4-bit quantization.

\subsubsection{COCO Val2017 Zero-Shot.}
COCO~\cite{COCO} contains 80 categories with scale-wise evaluation across small (AP$_s$), medium (AP$_m$), and large (AP$_l$), making it suitable for assessing scale robustness.
As shown in \cref{tab:coco-zeroshot}, PTQ again completely fails with AP 0.0, and QAT leaves a large gap from FP32.
QATMA consistently outperforms QAT, with improvements of +4.1 and +6.8 AP on YOLO-World-M and YOLO-World-L, and the gain further increases to +7.6 AP (40.9\%) on YOLO-World-X, again confirming larger gains at greater model scales.
These improvements also hold across all object scales, as illustrated by YOLO-World-M where AP$_s$, AP$_m$, and AP$_l$ improve by +3.2, +4.6, and +5.5, respectively.
The +7.4 gain in AP$_{75}$ on YOLO-World-L further demonstrates effective recovery of localization precision, confirming that QATMA restores not only classification alignment but also fine-grained localization capability.
QATMA similarly outperforms EMA+QC and QFD on COCO, as detailed in the supplementary material.

\subsection{Ablation Studies}\label{sec:ablation}

\subsubsection{Ablations on QATMA.}
\Cref{tab:ablation} analyzes the individual contributions of the curriculum strategy and KD on YOLO-World-L with 4-4-8 (Ch-T-H) quantization.
Applying only the curriculum (CQAT) to QAT (AP 12.8) yields AP 14.0 (+1.2), while applying KD alone yields only AP 13.1 (+0.3).
Combining both achieves AP 16.1 (+3.3), far exceeding the sum of their individual gains (+1.5).
This strong synergy confirms that the curriculum is an essential foundation for KD to operate effectively.

\subsubsection{Ablations on TPSD.}
\Cref{tab:trkd-ablation} decomposes TPSD into its two components: region-text alignment and region-region alignment.
Using only CQAT with backbone feature distillation (\ie, conventional KD without TPSD) yields AP 14.5, only +0.5 over CQAT alone (14.0).
Distilling only region-text alignment yields AP 15.0 (+0.5), while distilling only region-region alignment yields AP 15.4 (+0.9).
Combining both, full TPSD achieves the best AP of 16.1, a further +1.6 that far exceeds the +0.5 from conventional feature distillation. This confirms that preserving cross-modal and intra-modal alignment is critical to low-bit OVOD accuracy.

\begin{table}[tb]
    \centering
    \begin{minipage}[t]{0.48\columnwidth}
        \centering
        \caption{Ablations on QATMA. KD includes feature distillation and TPSD. LVIS zero-shot AP is reported.}
        \label{tab:ablation}
        \begin{tabular}{@{}ccc@{}}
            \toprule
            Curriculum & KD & AP \\
            \midrule
            \xmark & \xmark & 12.8 \\
            \cmark & \xmark & 14.0 \\
            \xmark & \cmark & 13.1 \\
            \midrule
            \cmark & \cmark & \textbf{16.1} \\
            \bottomrule
        \end{tabular}
    \end{minipage}\hfill
    \begin{minipage}[t]{0.48\columnwidth}
        \centering
        \caption{TPSD component analysis. All variants use CQAT with feature distillation. LVIS zero-shot AP is reported.}
        \label{tab:trkd-ablation}
        \begin{tabular}{@{}lc@{}}
          \toprule
          TPSD variant & AP \\
          \midrule
          w/o TPSD (feat only) & 14.5 \\
          region-text & 15.0 \\
          region-region & 15.4 \\
          \midrule
          full TPSD & \textbf{16.1} \\
          \bottomrule
        \end{tabular}
    \end{minipage}
\end{table}

\subsubsection{Effect of Curriculum Stages.}
\begin{figure}[tb]
    \centering
    \begin{minipage}[c]{0.48\columnwidth}
        \centering
        \captionof{table}{Effect of curriculum stages. 3-stage uses 1/3 data per stage. LVIS zero-shot AP is reported.}
        \label{tab:curriculum-stage}
        \begin{tabular}{@{}lcc@{}}
          \toprule
          Curriculum stages & KD & AP \\
          \midrule
          \multirow{2}{*}{2 (backbone$\rightarrow$neck-head)}
            & \xmark & 14.0 \\
            & \cmark & \textbf{16.1} \\
          \midrule
          \multirow{2}{*}{3 (backbone$\rightarrow$neck$\rightarrow$head)}
            & \xmark & 11.0 \\
            & \cmark & 15.3 \\
          \bottomrule
        \end{tabular}
        \medskip
        \captionof{table}{Generalization of QATMA to OmDet-Turbo~\cite{OmDetTurbo} under 4-4-8 (Ch-T-H) quantization. LVIS zero-shot AP is reported.}
        \label{tab:omdet-generalization}
        \begin{tabular}{@{}lcccc@{}}
          \toprule
          Method & AP & AP$_r$ & AP$_c$ & AP$_f$ \\
          \midrule
          FP32 & 29.1 & 22.5 & 28.2 & 31.1 \\
          \midrule
          QAT & 12.8 & 9.0 & 11.1 & 14.9 \\
          Ours & \textbf{17.6} & \textbf{10.9} & \textbf{16.9} & \textbf{19.5} \\
          \bottomrule
        \end{tabular}
    \end{minipage}\hfill
    \begin{minipage}[c]{0.48\columnwidth}
        \centering
        \includegraphics[width=\columnwidth]{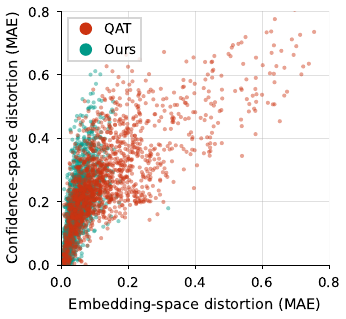}
        \captionof{figure}{Embedding-space vs.\ confidence-space distortion (MAE of the region-region similarity relative to FP32). Each point is an (image, category) group with $\geq$10 anchors.}
        \label{fig:error-scatter}
    \end{minipage}
\end{figure}

\Cref{tab:curriculum-stage} compares two-stage (backbone$\rightarrow$neck-head) and three-stage (backbone$\rightarrow$neck$\rightarrow$head) curriculum strategies.
Two-stage is consistently superior, both without KD (14.0 vs.\ 11.0) and with KD (16.1 vs.\ 15.3). The neck and head jointly perform cross-modal fusion and region-text matching, forming a natural functional unit. Notably, our KD improves both partitions substantially (+2.1 for two-stage and +4.3 for three-stage) and narrows their gap from 3.0 to 0.8. This demonstrates that our KD is robust to the choice of curriculum partition.

\subsection{Analysis}\label{sec:analysis}

\subsubsection{Architecture Generality.}

We apply QATMA to OmDet-Turbo~\cite{OmDetTurbo}\footnote{It releases only checkpoints; we re-implement its training and evaluation pipeline.}, a real-time OVOD detector whose DETR-family architecture is fundamentally distinct from the one-stage anchor-free YOLO-World. TPSD is applied identically, except that positive samples are the top-$K$ decoder queries per instance rather than dense TAL anchors. \Cref{tab:omdet-generalization} reports zero-shot results on LVIS \texttt{minival} under 4-4-8 (Ch-T-H) quantization after one epoch of QAT. QATMA improves over QAT by $+4.8$ AP, confirming generalization across heterogeneous OVOD architectures.

\subsubsection{Embedding-to-Confidence Alignment Transfer.}
To verify that preserving intra-modal alignment also benefits the predictions, we measure how quantization distorts the region-region pairwise similarity in two spaces on 512 LVIS samples (\cref{fig:error-scatter}): among region embeddings and among their confidence score vectors. Preserving the latter is known to retain a model's discriminative knowledge and output-distribution fidelity~\cite{CKA}. The two distortions are strongly correlated (Spearman correlation of $0.76$), showing that embedding-space distortion propagates directly to the predictions. QATMA reduces both far more than QAT (MAE $0.1455\to0.0494$ and $0.2766\to0.1813$) because TPSD explicitly distills this alignment, which QAT's task-only objective leaves distorted.

\subsubsection{Qualitative Analysis.}
\begin{figure}[tb]
    \centering
    \newcommand{\colwidth}{0.22\textwidth}
    \begin{minipage}{\colwidth}\centering\textbf{FP32}\end{minipage}%
    \begin{minipage}{\colwidth}\centering\textbf{QAT}\end{minipage}%
    \begin{minipage}{\colwidth}\centering\textbf{Ours}\end{minipage}

    \includegraphics[width=\colwidth]{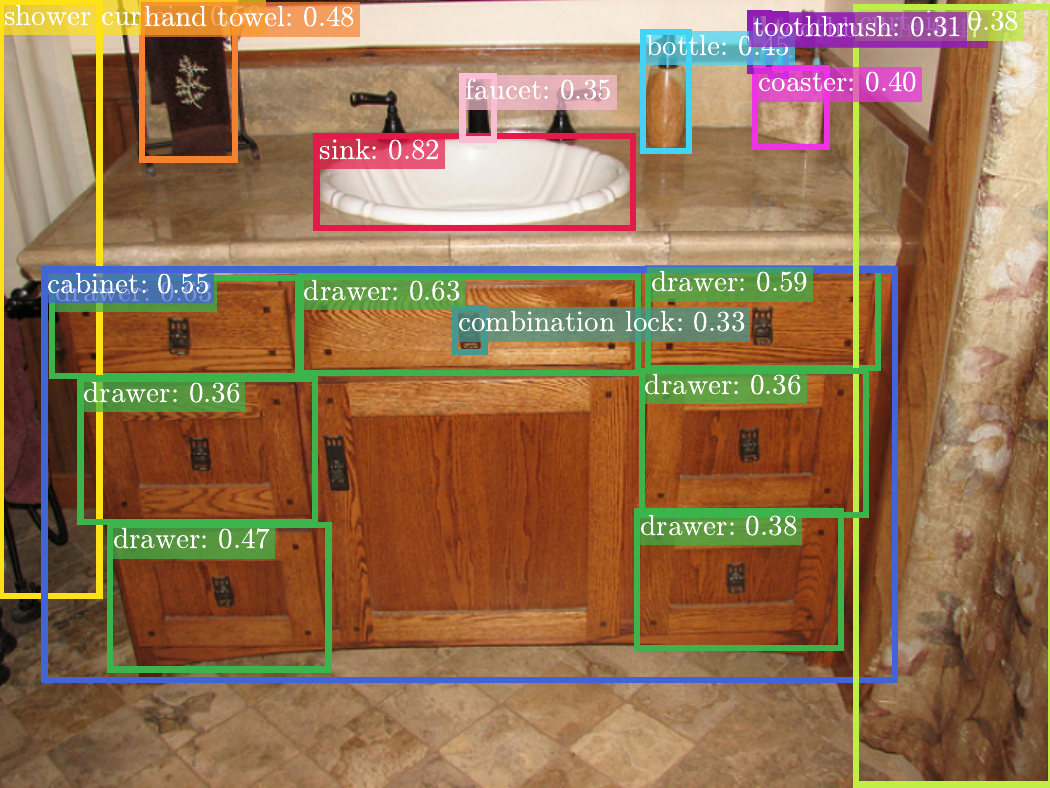}%
    \includegraphics[width=\colwidth]{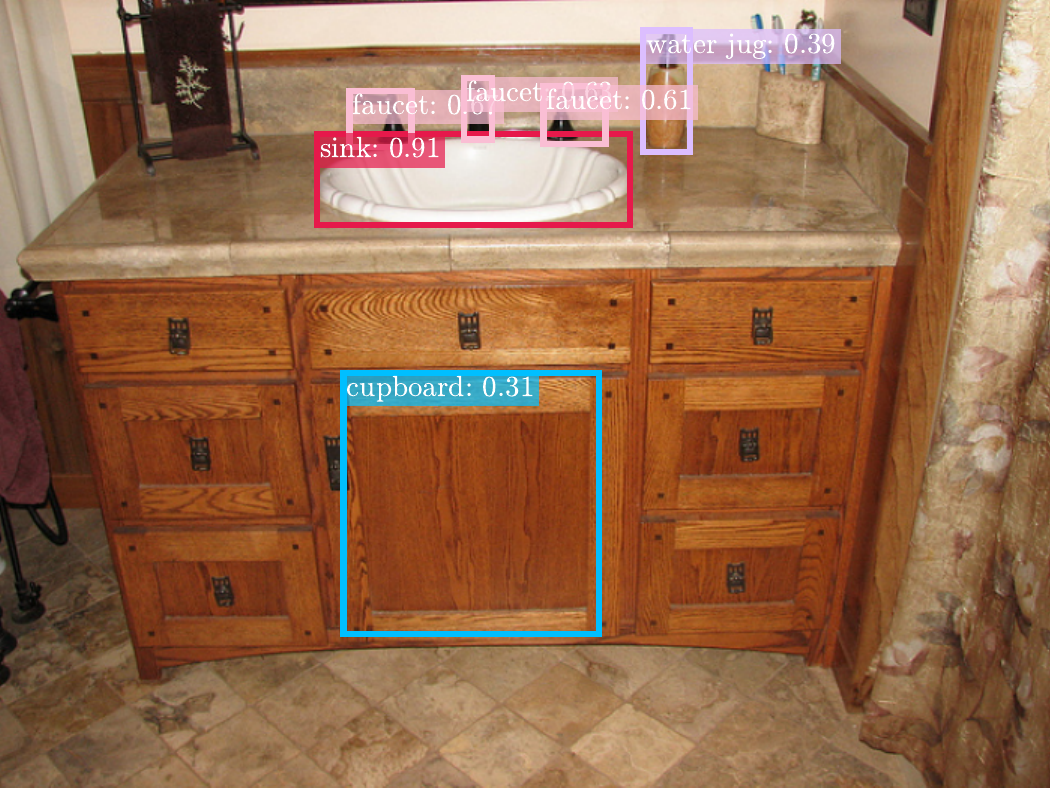}%
    \includegraphics[width=\colwidth]{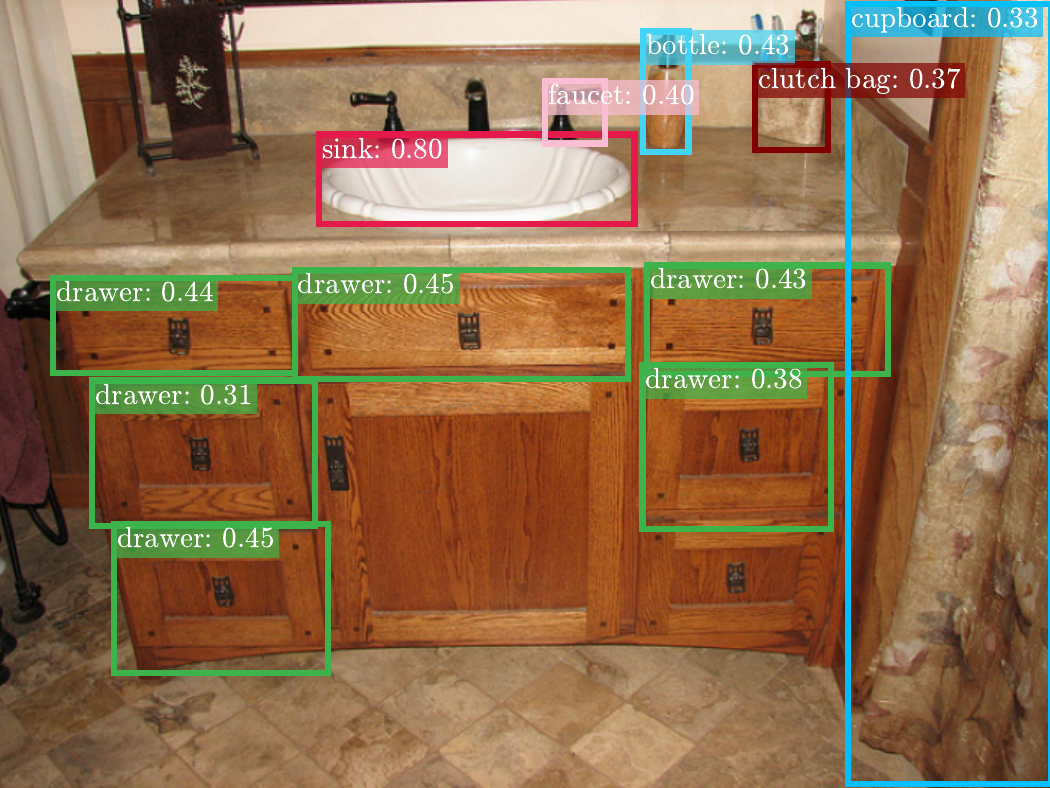}

    \includegraphics[width=\colwidth]{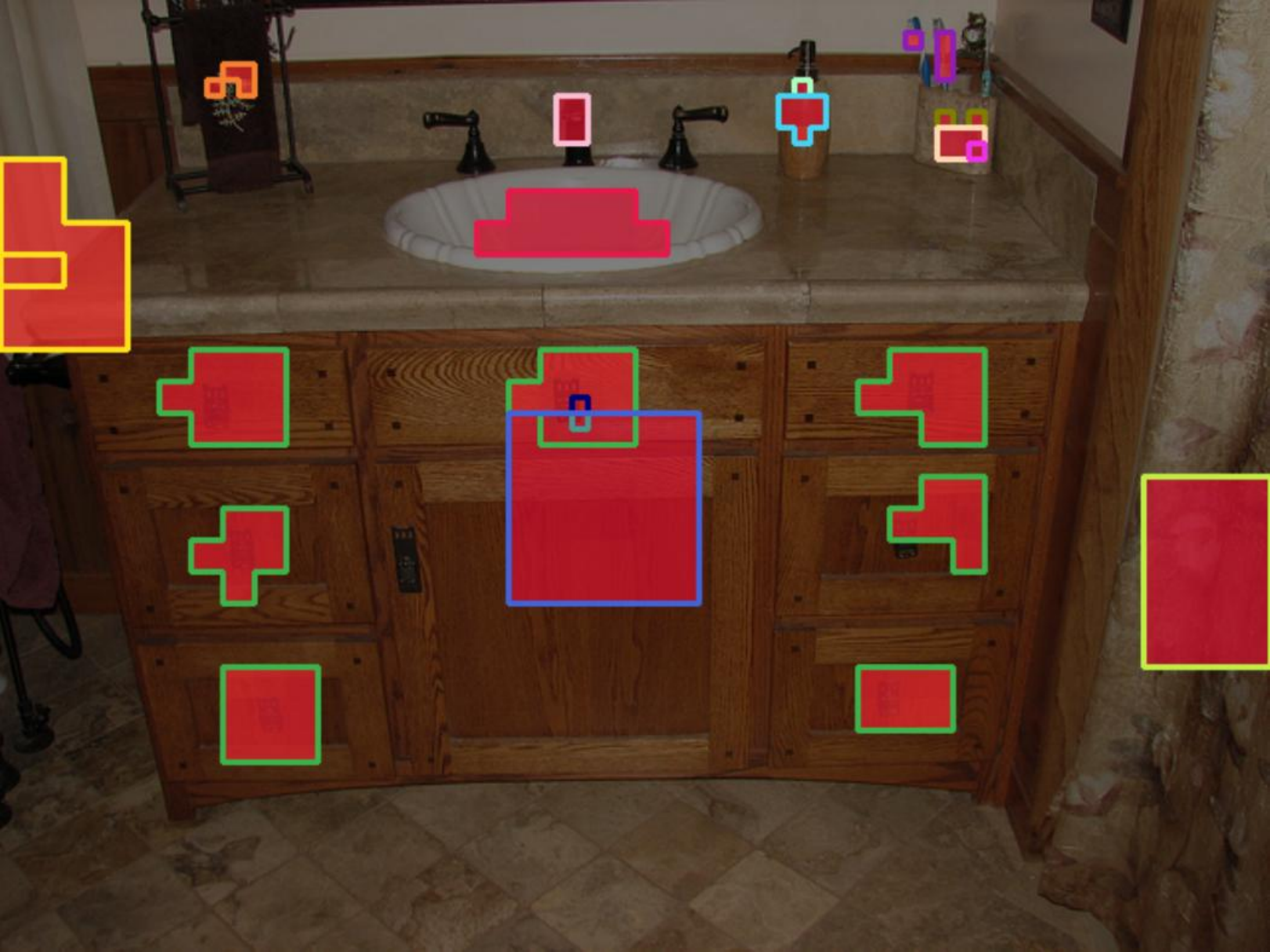}%
    \includegraphics[width=\colwidth]{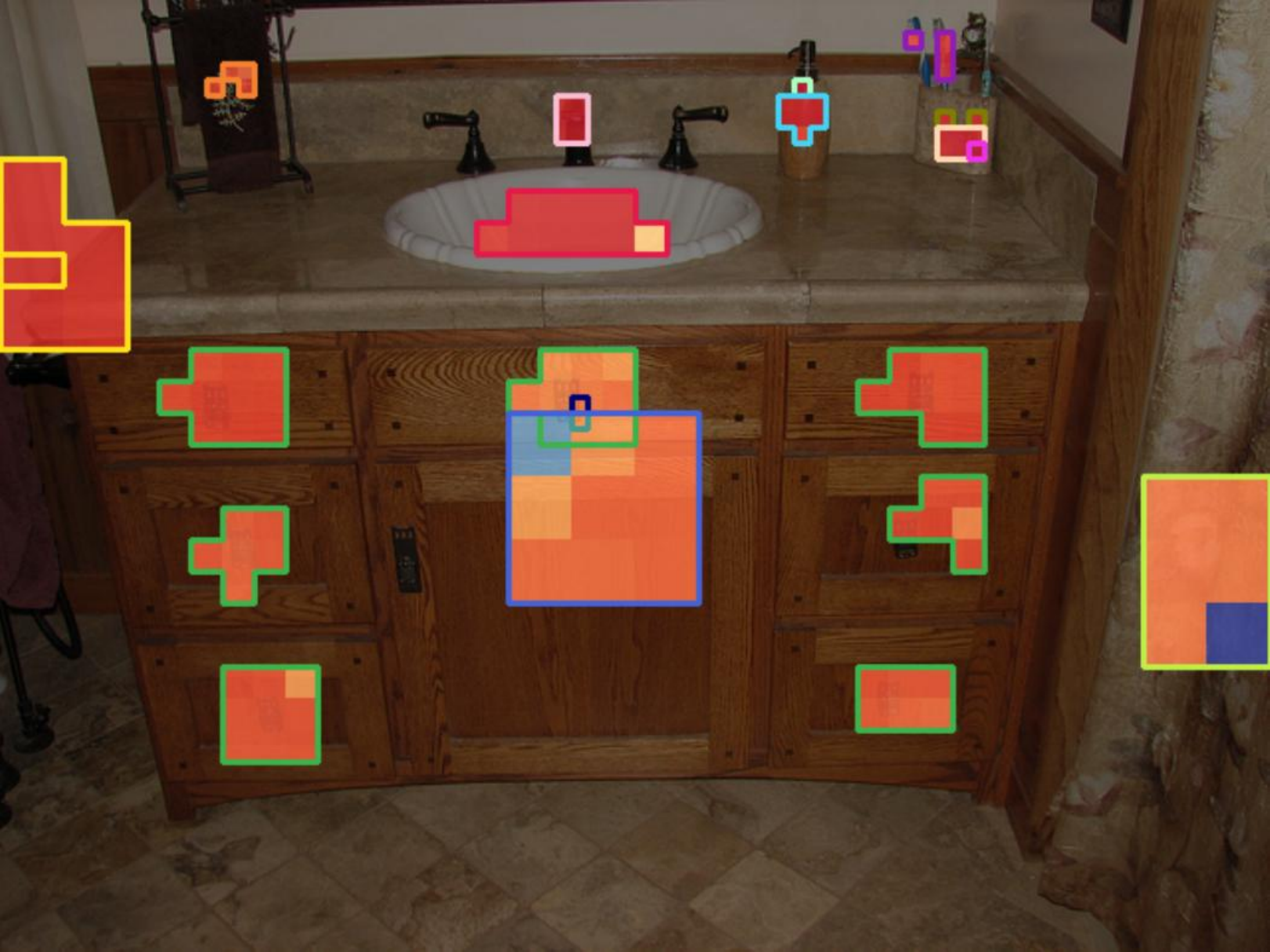}%
    \includegraphics[width=\colwidth]{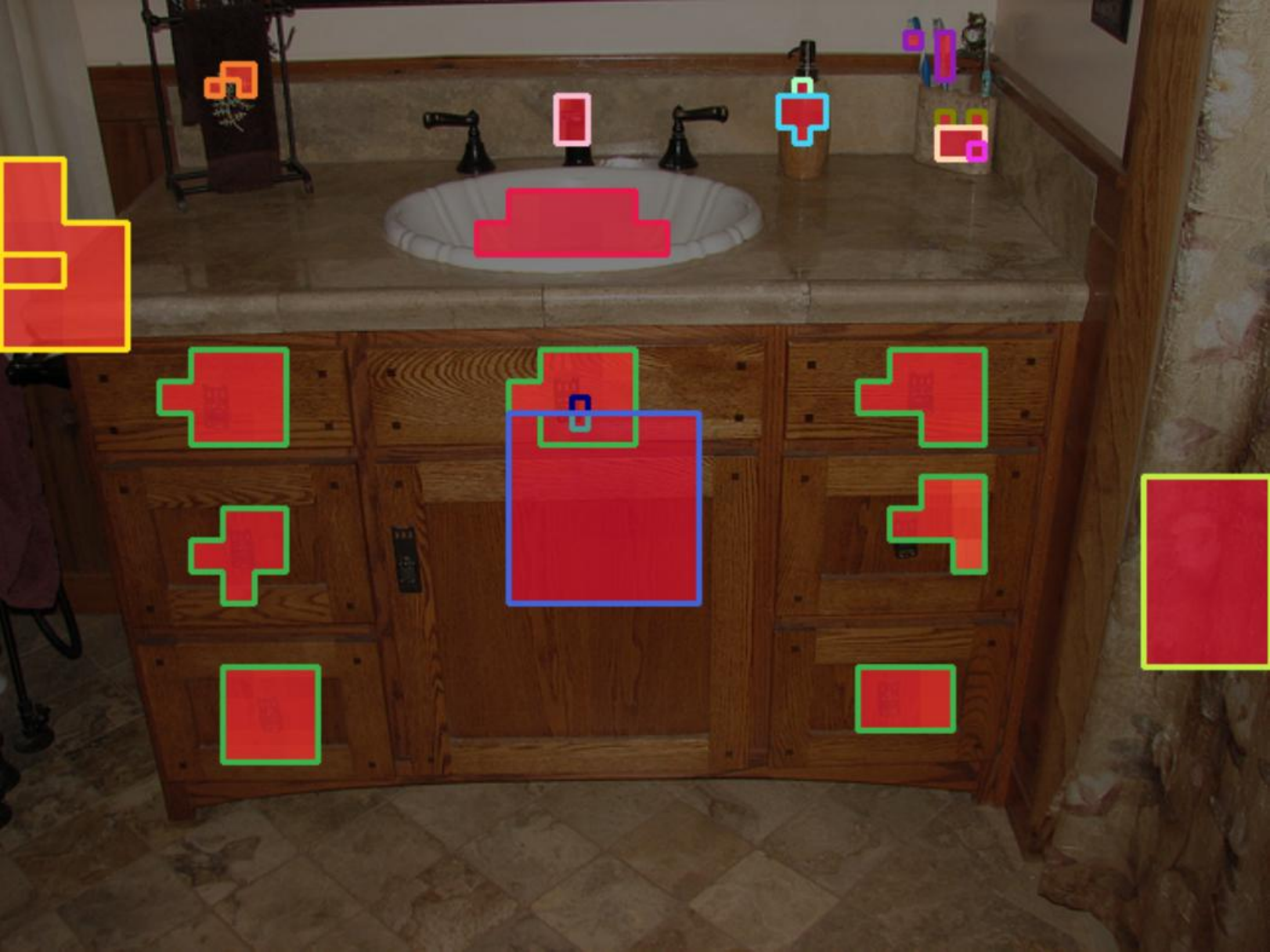}

    \includegraphics[width=0.4\textwidth]{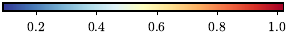}

    \caption{Qualitative comparison on YOLO-World-L (4-4-8, Ch-T-H). (\emph{Top}) Detection results. (\emph{Bottom}) Region-region similarity heatmap of average pairwise cosine similarity among same-class anchors. QAT distorts both detection and similarity patterns of FP32, whereas QATMA restores them.}
    \label{fig:qualitative}
\end{figure}

\Cref{fig:qualitative} presents a qualitative comparison of zero-shot detection results on an LVIS image.
In the detection results (top), QAT shows a significant reduction in the number of detections compared to FP32, missing fine-grained objects, including multiple drawers.
QATMA substantially recovers these detections.
For region-region similarity (bottom), we visualize the average pairwise cosine similarity of each anchor with other anchors of the same class as a heatmap.
We observe that QATMA exhibits similarity patterns close to those of FP32, whereas QAT displays notably different patterns, qualitatively supporting the effectiveness of QATMA in preserving region-region (intra-modal) alignment.

\section{Conclusion}
In this paper, we propose QATMA to address the severe degradation of cross-modal and intra-modal alignment inherent in extremely low-bit OVOD. To suppress the accumulation of quantization errors, we formulate CQAT, which provides a stable foundation for effective KD via stage-by-stage network partitioning. Concurrently, TPSD explicitly transfers the teacher's cross-modal and intra-modal alignment through text-anchored pairwise cosine similarities in the joint embedding space. Extensive evaluations on LVIS and COCO zero-shot benchmarks demonstrate that QATMA effectively restores the degraded alignment, significantly outperforming existing QAT baselines under extremely low-bit settings with improvements of up to 4.3 AP and 7.6 AP, respectively.

% ---- Bibliography ----
\bibliographystyle{splncs04}
\bibliography{main}

\ifincludesupp
  \clearpage
  \appendix
\setcounter{figure}{4}
\setcounter{table}{6}

\section{More Experimental Results}

\subsection{Full COCO Zero-Shot Comparison}

\Cref{tab:supp-coco} reports the complete COCO \texttt{val2017} zero-shot results, including the EMA+QC~\cite{OscQuant} and QFD~\cite{QFD} closed-vocabulary detection QAT baselines omitted from the main paper for brevity. Across all model scales and object sizes, QATMA consistently outperforms both baselines as well as naive QAT (LSQ~\cite{LSQ,LSQPlus}), whereas EMA+QC and QFD perform comparably to naive QAT.

Under our zero-shot protocol, each category is recognized by aligning its text query with region embeddings in the joint embedding space. The detector is trained on Objects365v2~\cite{Objects365} and GQA~\cite{GQA}, never on COCO. EMA+QC and QFD, by contrast, were validated for closed-vocabulary detection QAT---even on COCO---yet neither targets this alignment. EMA+QC mitigates weight and activation oscillations during low-bit QAT, a modality-agnostic stability mechanism. QFD distills only the teacher's FPN-level feature maps, a visual-feature-level supervision that lies outside the joint embedding space. Both therefore leave the cross-modal and intra-modal alignment distorted and degrade even on COCO. While effective in the closed-vocabulary setting they were designed for, these techniques do not extend to open-vocabulary detection.

\begin{table}[!ht]
    \caption{Full COCO \texttt{val2017} zero-shot comparison, including the EMA+QC~\cite{OscQuant} and QFD~\cite{QFD} closed-vocabulary detection QAT baselines (4-4-8, Ch-T-H granularity).}
    \label{tab:supp-coco}
    \centering
    \begin{tabular*}{\columnwidth}{@{\extracolsep{\fill}}llclcccccc@{}}
      \toprule
      Model & Bits & BOPs(T) & Method & AP & AP$_{50}$ & AP$_{75}$ & AP$_s$ & AP$_m$ & AP$_l$ \\
      \midrule
      \multirow{5}{*}{YOLO-World-M}
        & FP32 & 44.56 & - & 41.9 & 57.0 & 45.5 & 26.7 & 46.1 & 55.0 \\
      \cmidrule(l){2-10}
        & \multirow{4}{*}{4-4-8} & \multirow{4}{*}{2.05} & QAT & 22.0 & 32.1 & 23.6 & 7.8 & 23.5 & 32.4 \\
        & & & EMA+QC~\cite{OscQuant} & 21.2 & 31.1 & 22.7 & 8.3 & 22.7 & 31.3 \\
        & & & QFD~\cite{QFD} & 21.6 & 31.4 & 23.3 & 7.9 & 23.7 & 31.8 \\
        & & & Ours & \textbf{26.1} & \textbf{38.1} & \textbf{28.1} & \textbf{11.0} & \textbf{28.1} & \textbf{37.9} \\
      \midrule
      \multirow{5}{*}{YOLO-World-L}
        & FP32 & 90.60 & - & 45.1 & 60.7 & 48.9 & 29.8 & 49.8 & 57.5 \\
      \cmidrule(l){2-10}
        & \multirow{4}{*}{4-4-8} & \multirow{4}{*}{3.09} & QAT & 18.4 & 27.2 & 19.6 & 7.5 & 20.1 & 26.9 \\
        & & & EMA+QC~\cite{OscQuant} & 18.4 & 27.2 & 19.6 & 7.6 & 20.1 & 27.2 \\
        & & & QFD~\cite{QFD} & 18.2 & 26.8 & 19.2 & 7.0 & 19.6 & 27.0 \\
        & & & Ours & \textbf{25.2} & \textbf{36.7} & \textbf{27.0} & \textbf{10.5} & \textbf{27.5} & \textbf{37.7} \\
      \midrule
      \multirow{5}{*}{YOLO-World-X}
        & FP32 & 140.66 & - & 46.7 & 62.8 & 50.9 & 31.8 & 51.2 & 60.8 \\
      \cmidrule(l){2-10}
        & \multirow{4}{*}{4-4-8} & \multirow{4}{*}{4.21} & QAT & 18.6 & 26.8 & 19.9 & 7.8 & 20.5 & 27.0 \\
        & & & EMA+QC~\cite{OscQuant} & 17.7 & 25.8 & 19.0 & 7.7 & 18.6 & 25.6 \\
        & & & QFD~\cite{QFD} & 19.2 & 27.9 & 20.8 & 8.8 & 21.5 & 28.0 \\
        & & & Ours & \textbf{26.2} & \textbf{38.2} & \textbf{28.1} & \textbf{12.2} & \textbf{28.7} & \textbf{37.2} \\
      \bottomrule
    \end{tabular*}
\end{table}

\subsection{Effect of Quantization Granularity}

\Cref{tab:granularity-analysis} reports LVIS and COCO zero-shot performance on YOLO-World-L across different quantization granularities and bit-widths. Relaxing the activation bit-width to 5-bit (Ch-T-H 4-5-8) raises QAT AP to 23.0 on LVIS and 36.2 on COCO, and QATMA surpasses both, reaching 23.8 and 37.6, respectively. With per-channel activation granularity (Ch-Ch-H), even the extreme 3-bit setting (3-3-8) remains feasible, where QATMA still outperforms QAT on both LVIS (20.0 vs.\ 18.5) and COCO (32.6 vs.\ 30.7). These results confirm that QATMA is consistently robust with respect to granularity and bit-width changes on both benchmarks.

\begin{table}[!ht]
    \caption{Effect of quantization granularity and bit-width on YOLO-World-L. Ch-Ch-H = per-channel activations. LVIS and COCO zero-shot AP is reported.}
    \label{tab:granularity-analysis}
    \centering
    \begin{tabular}{@{}llccc@{}}
      \toprule
      Granularity & Bits & Method & LVIS & COCO \\
      \midrule
      \multirow{2}{*}{Ch-T-H}
        & \multirow{2}{*}{4-5-8} & QAT & 23.0 & 36.2 \\
        & & Ours & \textbf{23.8} & \textbf{37.6} \\
      \midrule
      \multirow{2}{*}{Ch-Ch-H}
        & \multirow{2}{*}{3-3-8} & QAT & 18.5 & 30.7 \\
        & & Ours & \textbf{20.0} & \textbf{32.6} \\
      \bottomrule
    \end{tabular}
\end{table}

\subsection{Data Split}

A natural question is why Curriculum QAT allocates only $1/3$ of the training data to Stage~1 (backbone) and the remaining $2/3$ to Stage~2 (neck-head). This split follows from the differing roles of the two modules. The backbone is a task-agnostic extractor of generic image features, whereas the neck-head is the task-relevant module that performs the fine-grained vision-language alignment---through cross-modal attention and region-text matching---and thus dominates the final detection accuracy. Beyond its functional role, the neck-head also contains more parameters than the backbone (27.7M versus 19.8M in YOLO-World-L, excluding the text encoder), further motivating a larger training budget for Stage~2. We therefore avoid over-investing in backbone stabilization: Stage~1 needs only to restore an adequate feature quality, so that the larger share of the data budget can be devoted to the harder and more critical recovery of the neck-head in Stage~2. \Cref{tab:data-split} supports this design: under identical conditions, an even $1/2 : 1/2$ split reaches only $15.0$ AP, whereas the proposed $1/3 : 2/3$ split attains $16.1$ AP in LVIS zero-shot detection. Allocating more data to Stage~2 is thus clearly beneficial.

\begin{table}[!ht]
    \caption{Effect of the curriculum data split between Stage~1 (backbone) and Stage~2 (neck-head) on YOLO-World-L (4-4-8, Ch-T-H). LVIS zero-shot AP is reported.}
    \label{tab:data-split}
    \centering
    \begin{tabular}{@{}ccc@{}}
      \toprule
      Stage~1 (backbone) & Stage~2 (neck-head) & AP \\
      \midrule
      1/3 & 2/3 & \textbf{16.1} \\
      1/2 & 1/2 & 15.0 \\
      \bottomrule
    \end{tabular}
\end{table}

\subsection{Hyperparameter Analysis}

To isolate the effect of TPSD, we adopt a head-only W4A4 setting: the backbone and neck remain in full precision, only the head is quantized to W4A4, and the whole network is trained. Quantizing the head alone is already severe: a simple MinMax PTQ on it collapses the LVIS zero-shot AP to $4.8$. We then study the sensitivity to the TPSD loss weight $\lambda_2$ on YOLO-World-L, trained on the Objects365v2~\cite{Objects365} validation split (80K images), where $\lambda_2 = 0$ reduces to naive QAT. As shown in \cref{tab:lambda-sensitivity}, TPSD improves over naive QAT across all $\lambda_2$, and the AP stays stable (within 0.4 among non-zero weights), confirming its robustness to this hyperparameter. PKD~\cite{PKD} is also known to be robust to its loss weight, so we set $\lambda_1 = \lambda_2 = 6$.

\begin{table}[!ht]
    \caption{TPSD loss weight $\lambda_2$ sensitivity on YOLO-World-L (head-only W4A4). LVIS zero-shot AP is reported.}
    \label{tab:lambda-sensitivity}
    \centering
    \begin{tabular}{@{}lcccccc@{}}
      \toprule
      $\lambda_2$ & 0 & 3 & 6 & 12 & 18 & 48 \\
      \midrule
      AP & 30.4 & 30.8 & \textbf{31.2} & \textbf{31.2} & 31.0 & 30.9 \\
      \bottomrule
    \end{tabular}
\end{table}

\subsection{Training Cost}

QATMA incurs no additional inference cost, since the curriculum and distillation affect only training. Its two-stage schedule also keeps the training overhead modest. As reported in \cref{tab:training-cost}, on YOLO-World-L QATMA increases the per-iteration compute from $27.1$k to $35.1$k GFLOPs ($+30\%$) and the total training time from $95.2$ to $112.4$ GPU-hours ($+18\%$) relative to plain QAT. This overhead stays small because Stage~1 is lightweight---the neck-head is frozen and the teacher is forwarded only through the backbone.

\begin{table}[!ht]
    \caption{Training cost on YOLO-World-L. QATMA adds no inference cost, and its two-stage training overhead over plain QAT is modest.}
    \label{tab:training-cost}
    \centering
    \begin{tabular}{@{}lcc@{}}
      \toprule
      Method & GFLOPs/iter & GPU-hours \\
      \midrule
      QAT & 27.1k & 95.2 \\
      Ours & 35.1k & 112.4 \\
      \bottomrule
    \end{tabular}
\end{table}

\section{Visualization}

\subsection{LVIS Zero-Shot Detection}

\Cref{fig:supp-lvis-vis} presents qualitative LVIS zero-shot detection results on YOLO-World-L (4-4-8, Ch-T-H). Naive QAT frequently misses or misclassifies objects. In the first row, for example, it labels a mouse (red) as a speaker (purple) and misses the laptop computer (green); in the second row it confuses a bathtub (yellow) with a sink (pink); and in the third row it classifies an elephant (red) as a horse (pink). Such errors stem from a failure to account for the context of neighboring regions, underscoring the importance of the cross-modal and intra-modal alignment. By preserving this alignment, QATMA corrects these mistakes and yields detections that closely match the FP32 reference.

\begin{figure}[p]
    \centering
    \newcommand{\vw}{0.25\textwidth}
    \setlength{\tabcolsep}{1.5pt}
    \begin{tabular}{@{}ccc@{}}
      \textbf{FP32} & \textbf{QAT} & \textbf{Ours} \\
      \includegraphics[width=\vw]{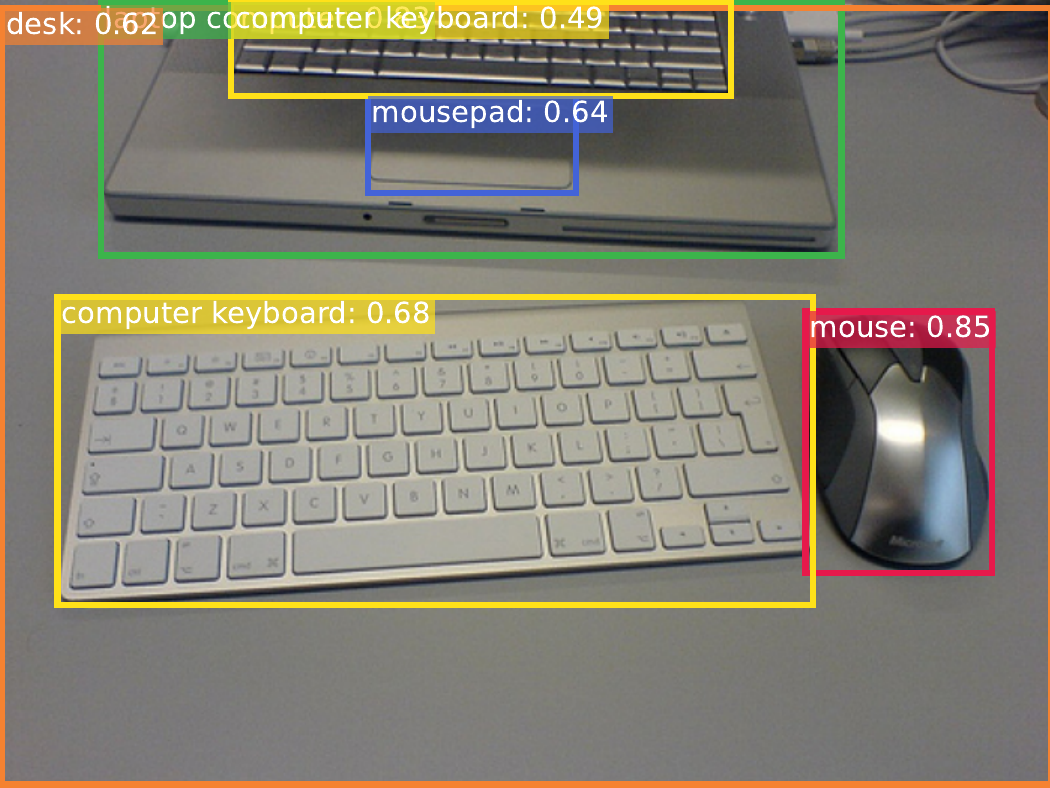} &
      \includegraphics[width=\vw]{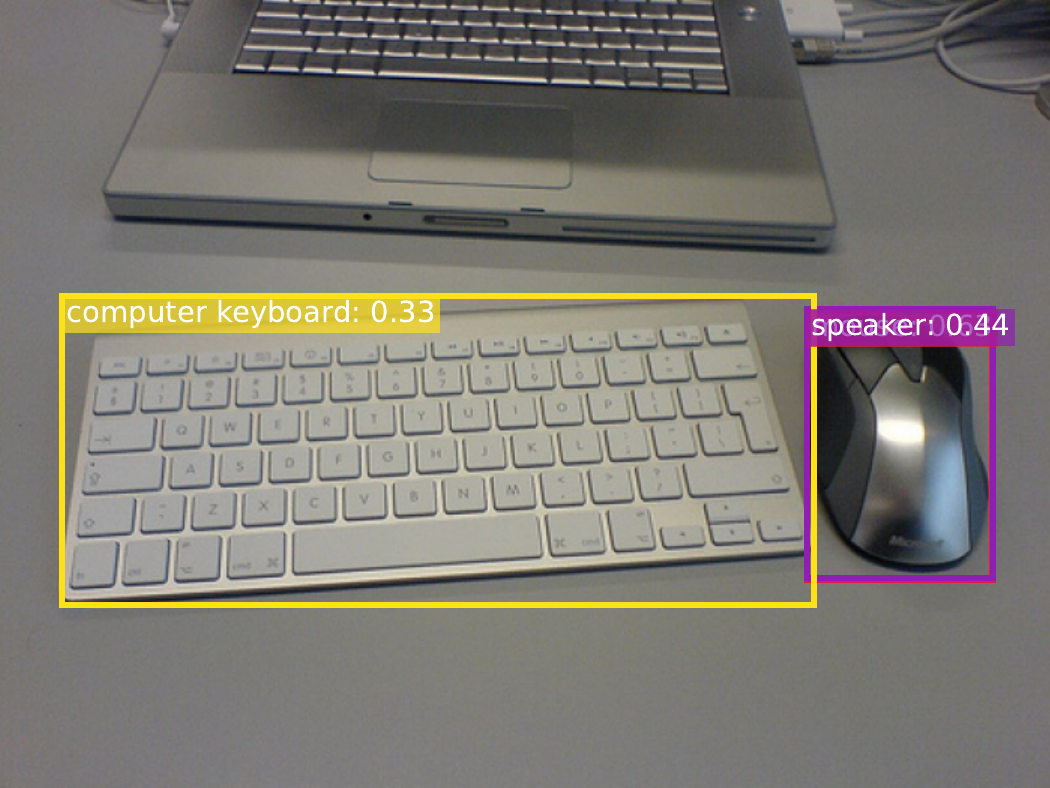} &
      \includegraphics[width=\vw]{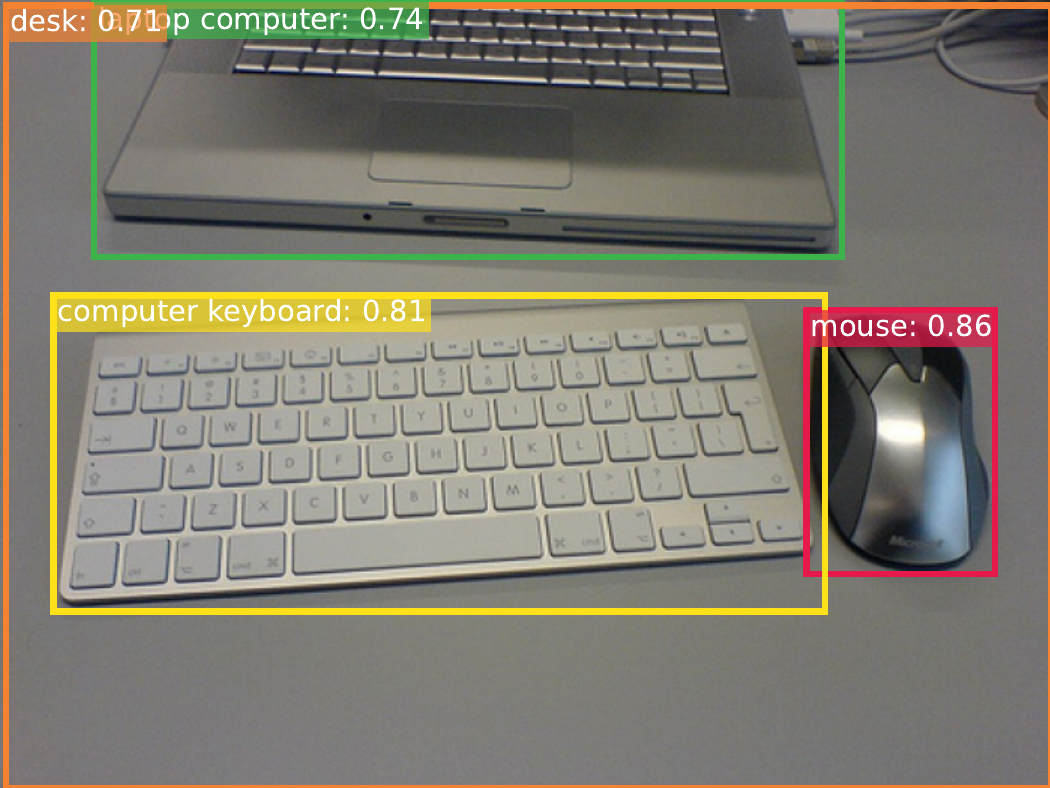} \\
      \includegraphics[width=\vw]{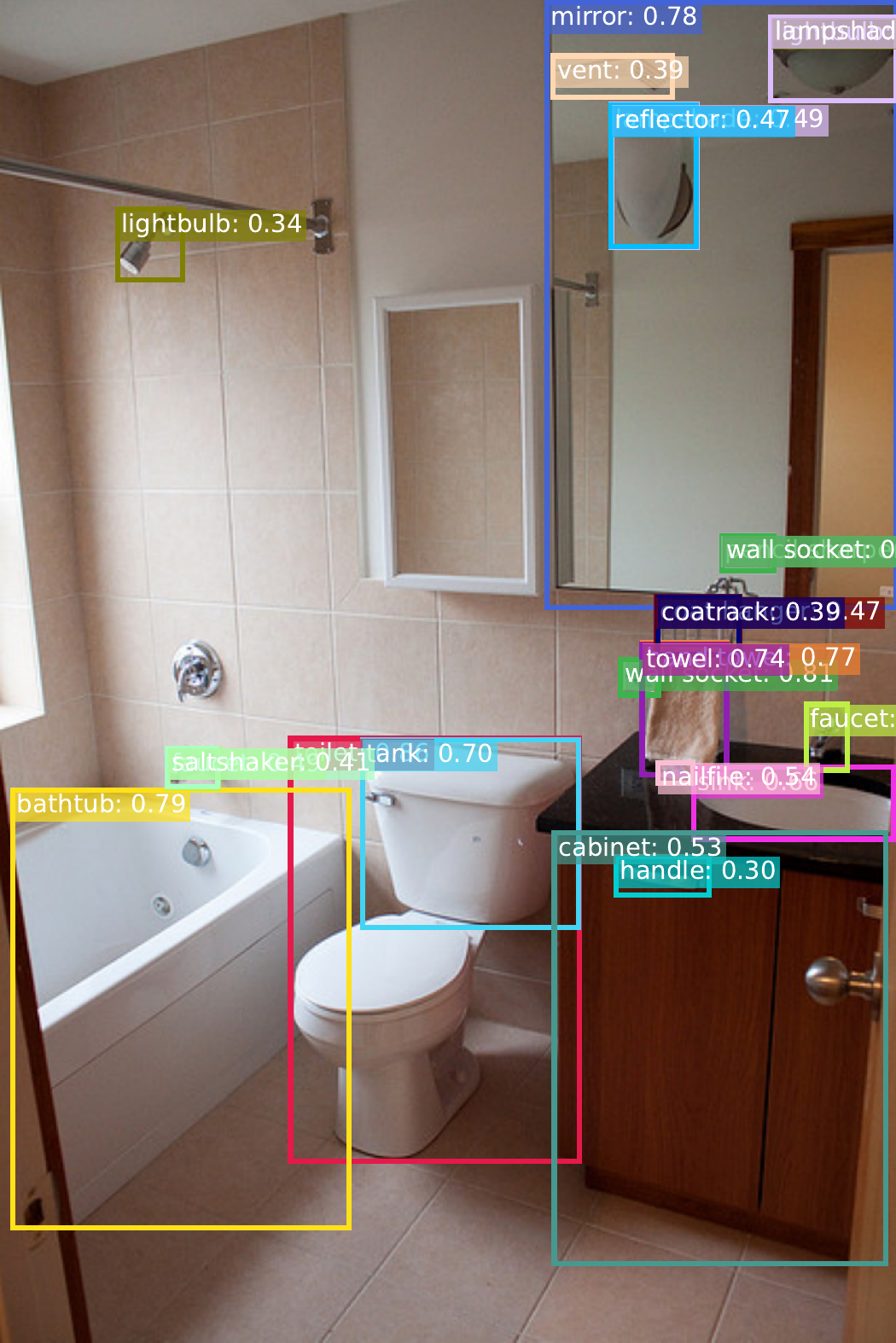} &
      \includegraphics[width=\vw]{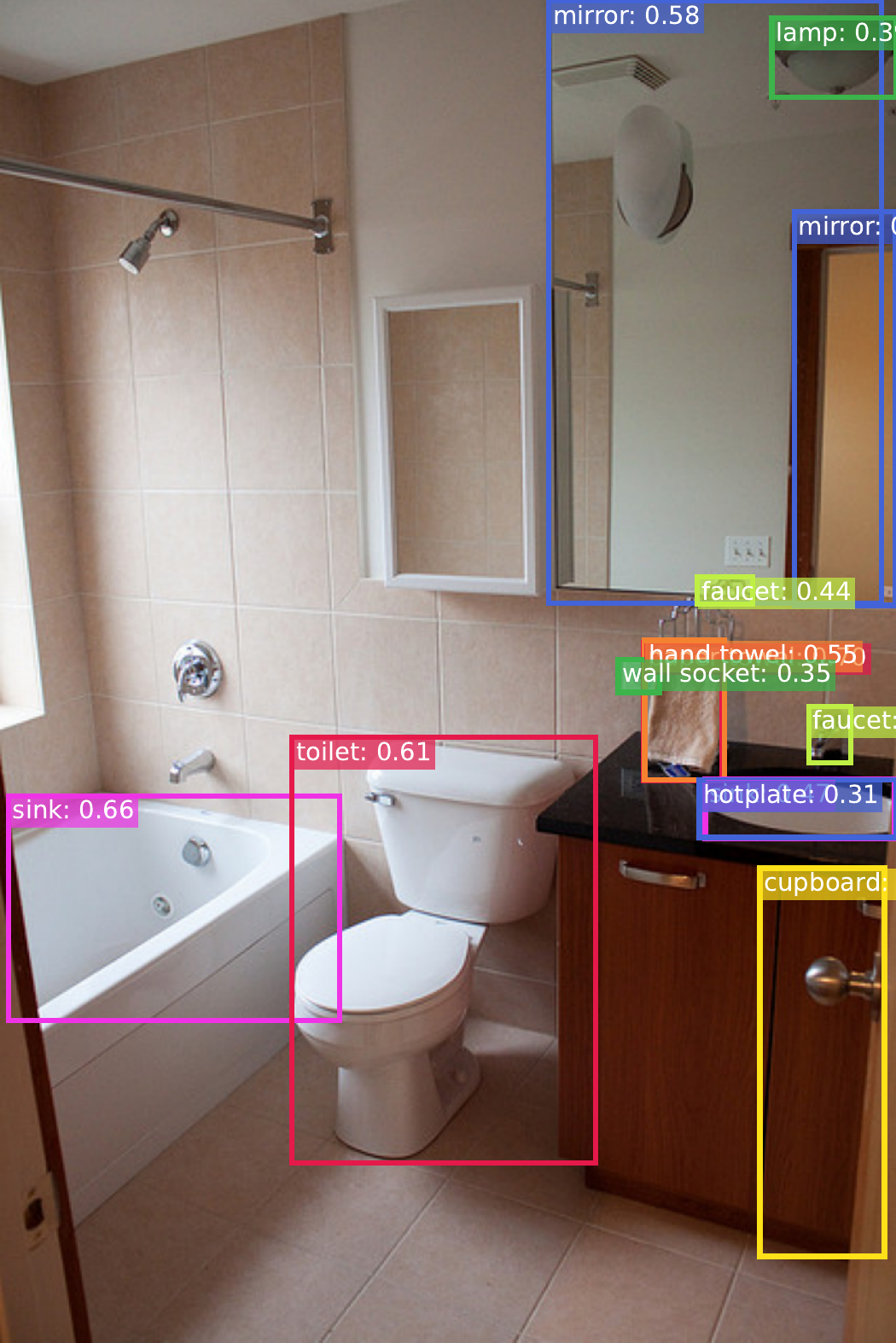} &
      \includegraphics[width=\vw]{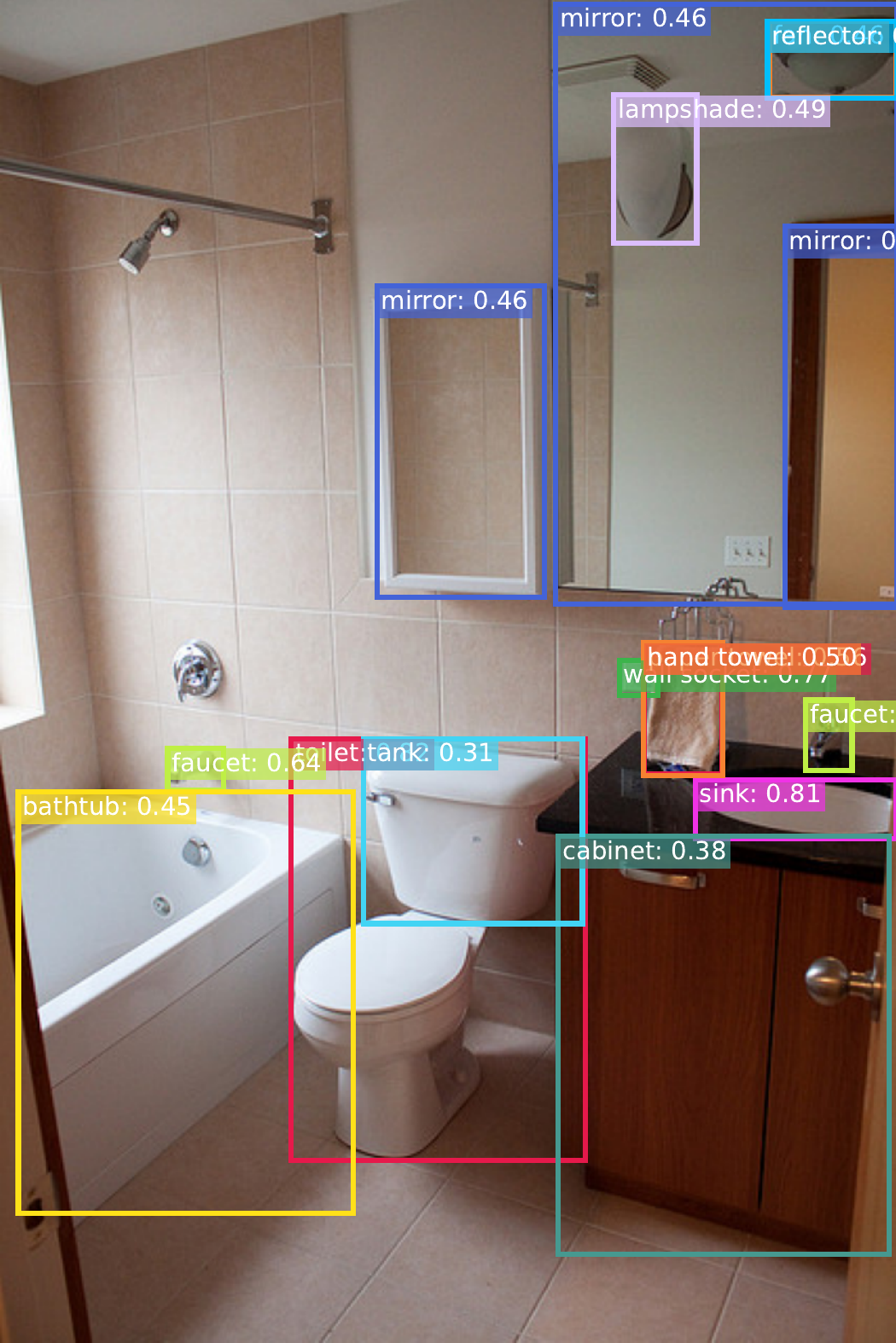} \\
      \includegraphics[width=\vw]{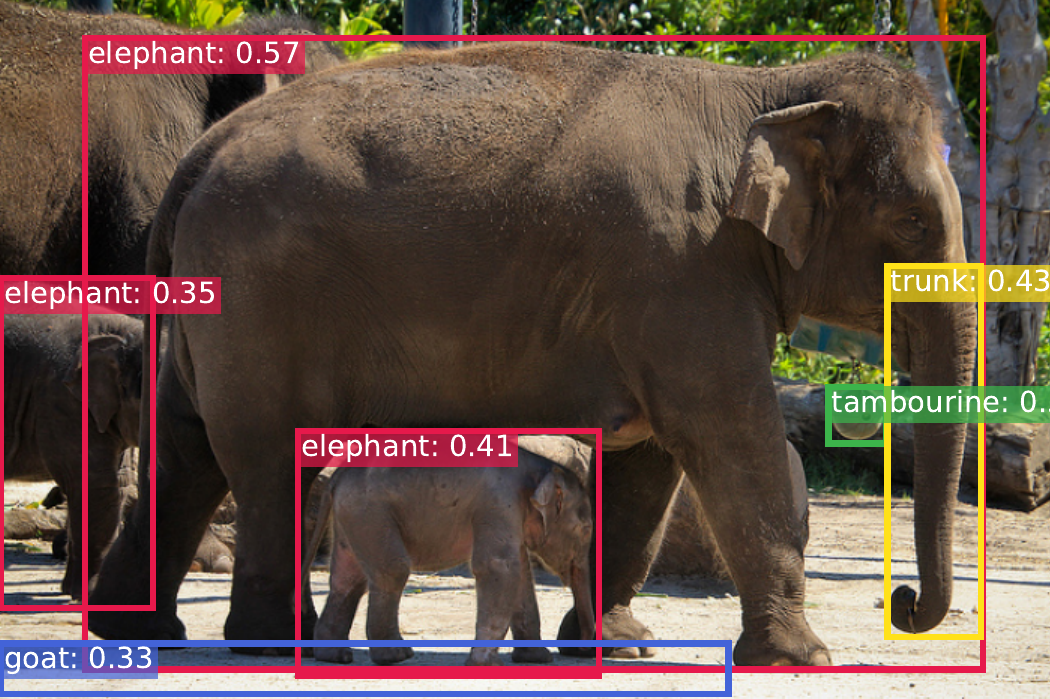} &
      \includegraphics[width=\vw]{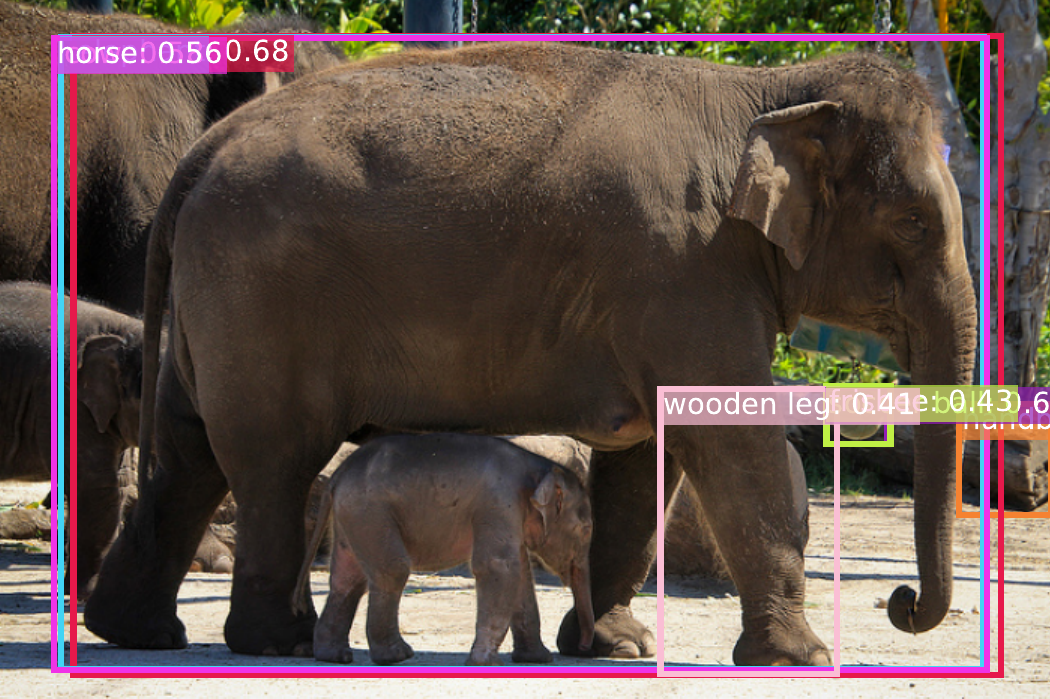} &
      \includegraphics[width=\vw]{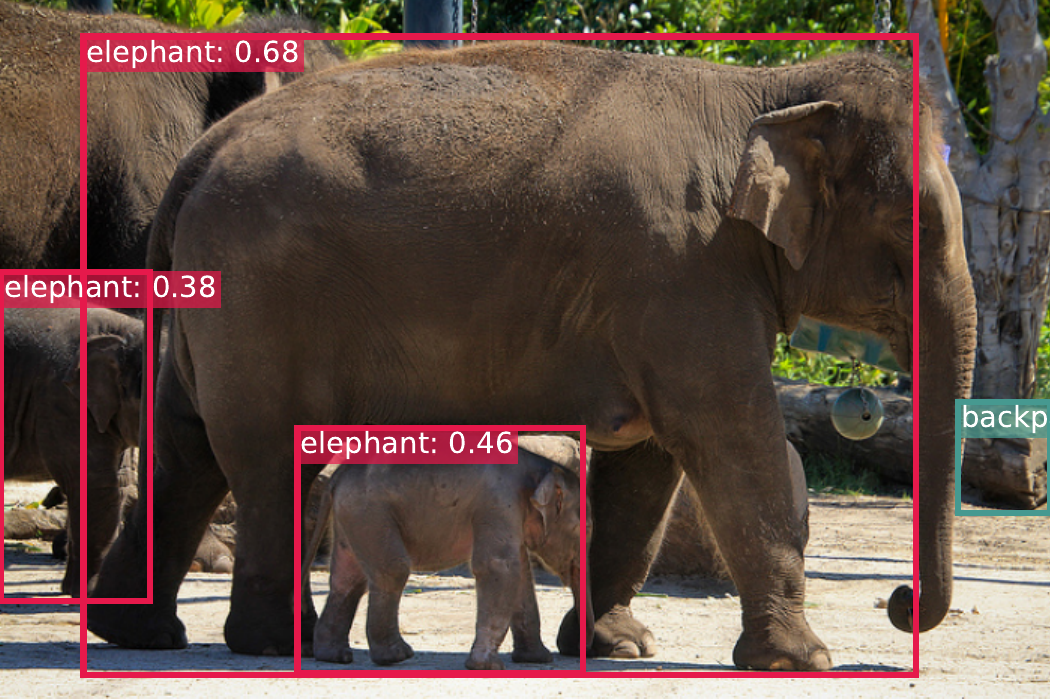} \\
      \includegraphics[width=\vw]{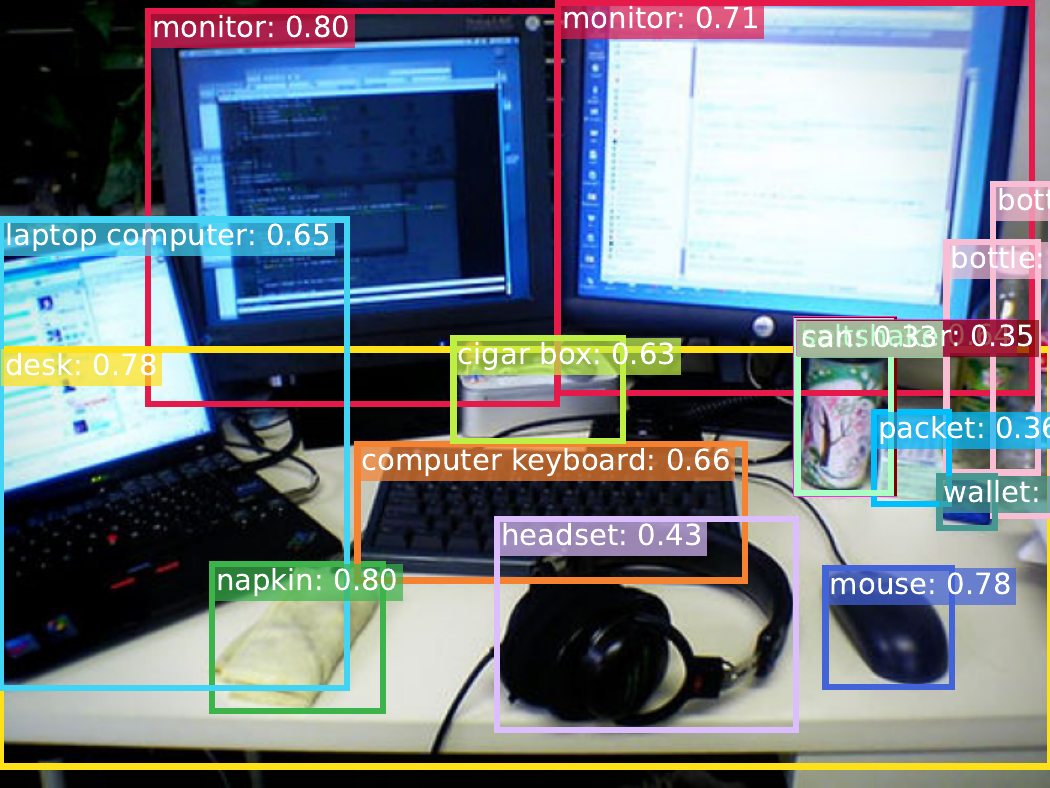} &
      \includegraphics[width=\vw]{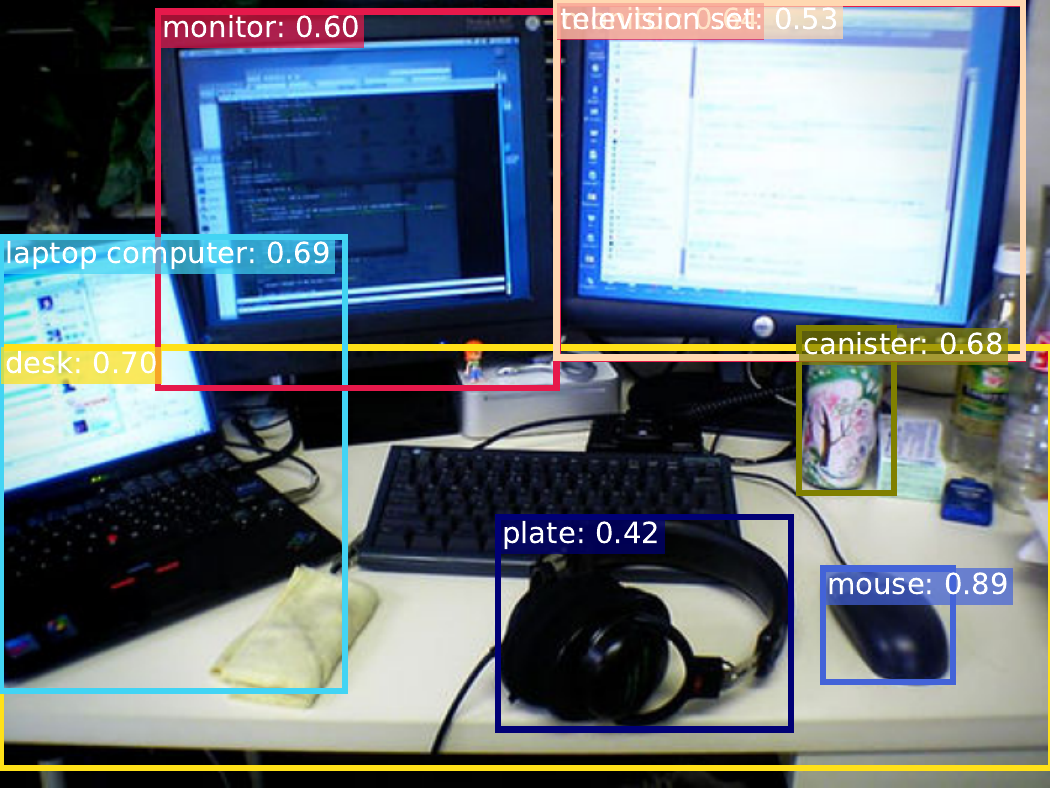} &
      \includegraphics[width=\vw]{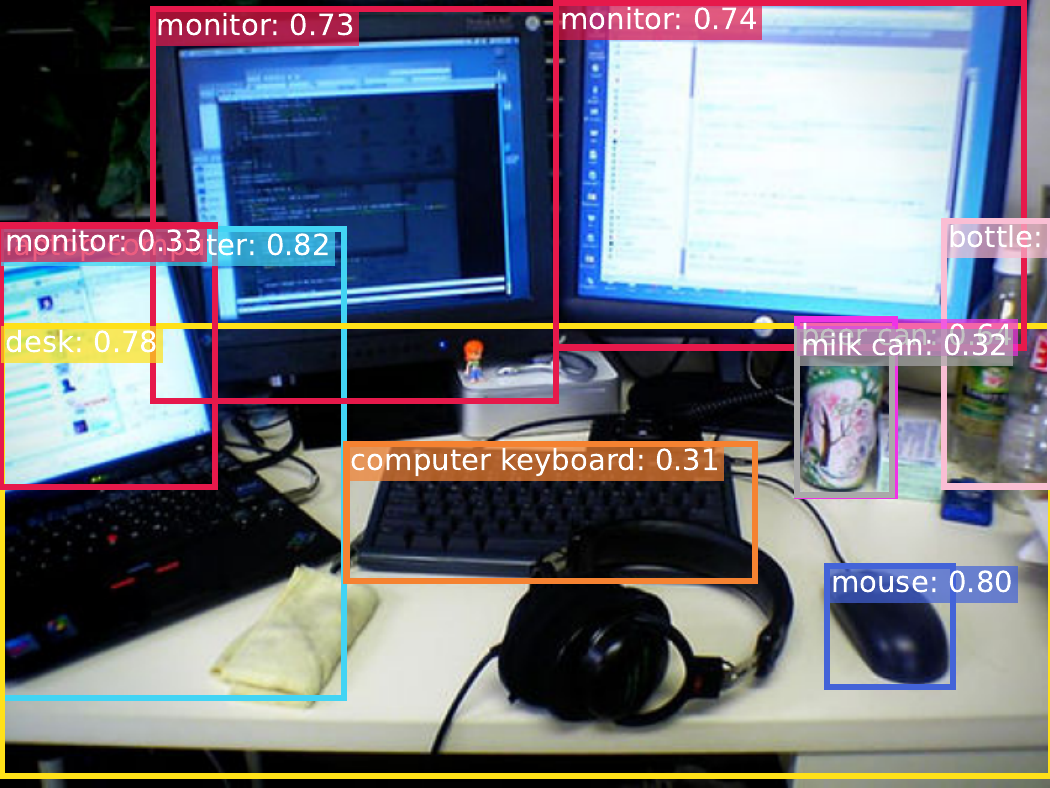} \\
      \includegraphics[width=\vw]{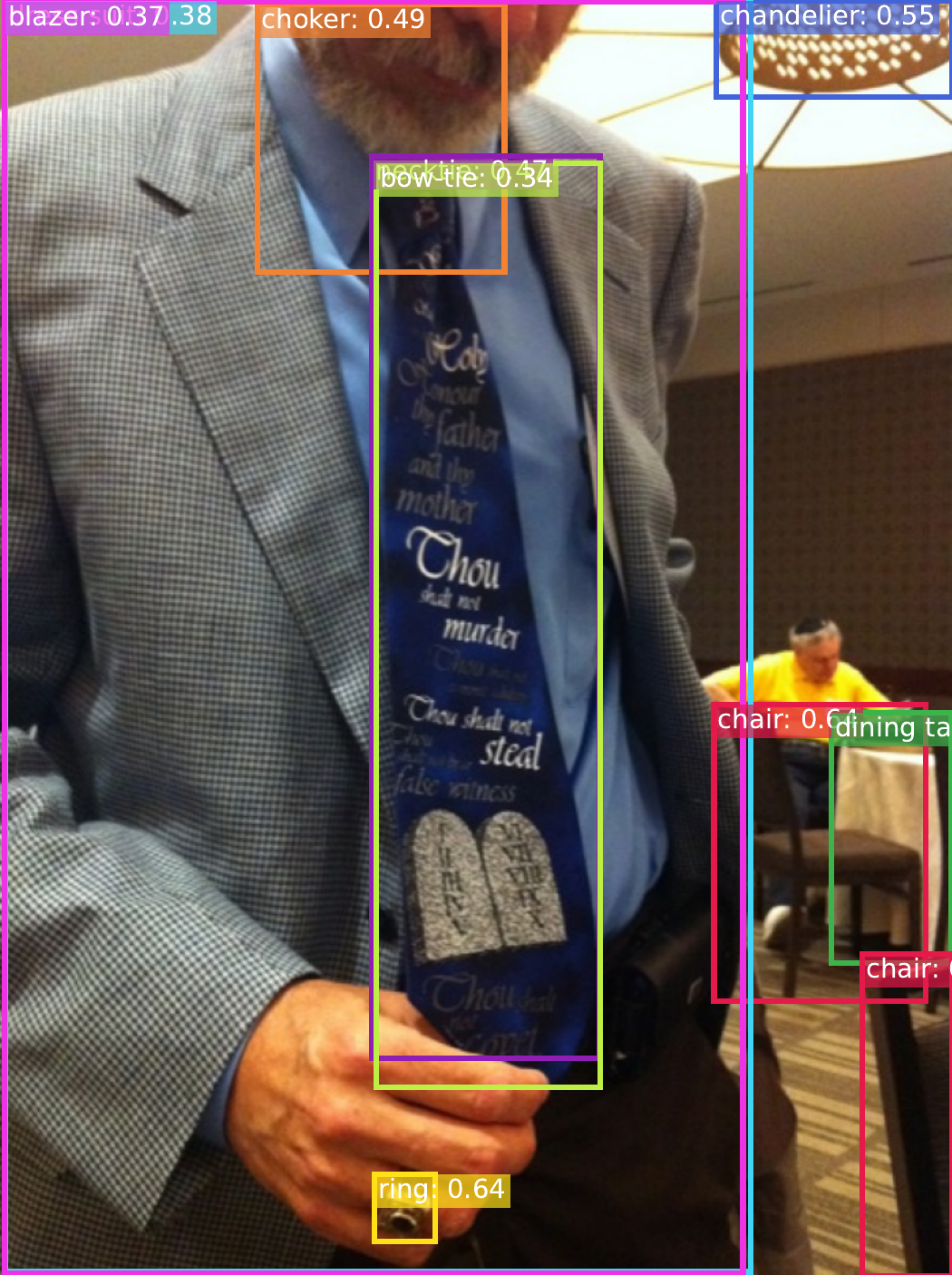} &
      \includegraphics[width=\vw]{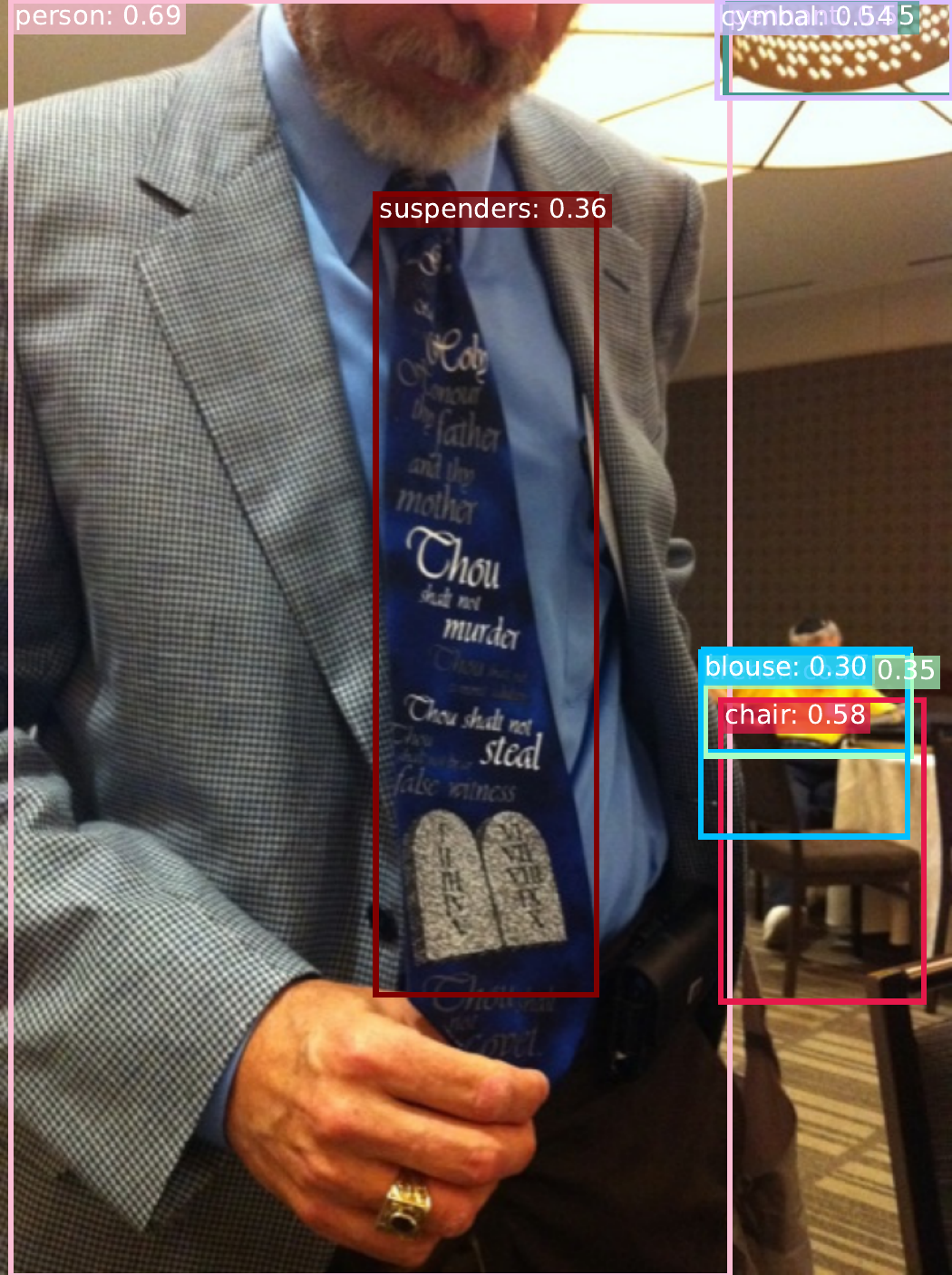} &
      \includegraphics[width=\vw]{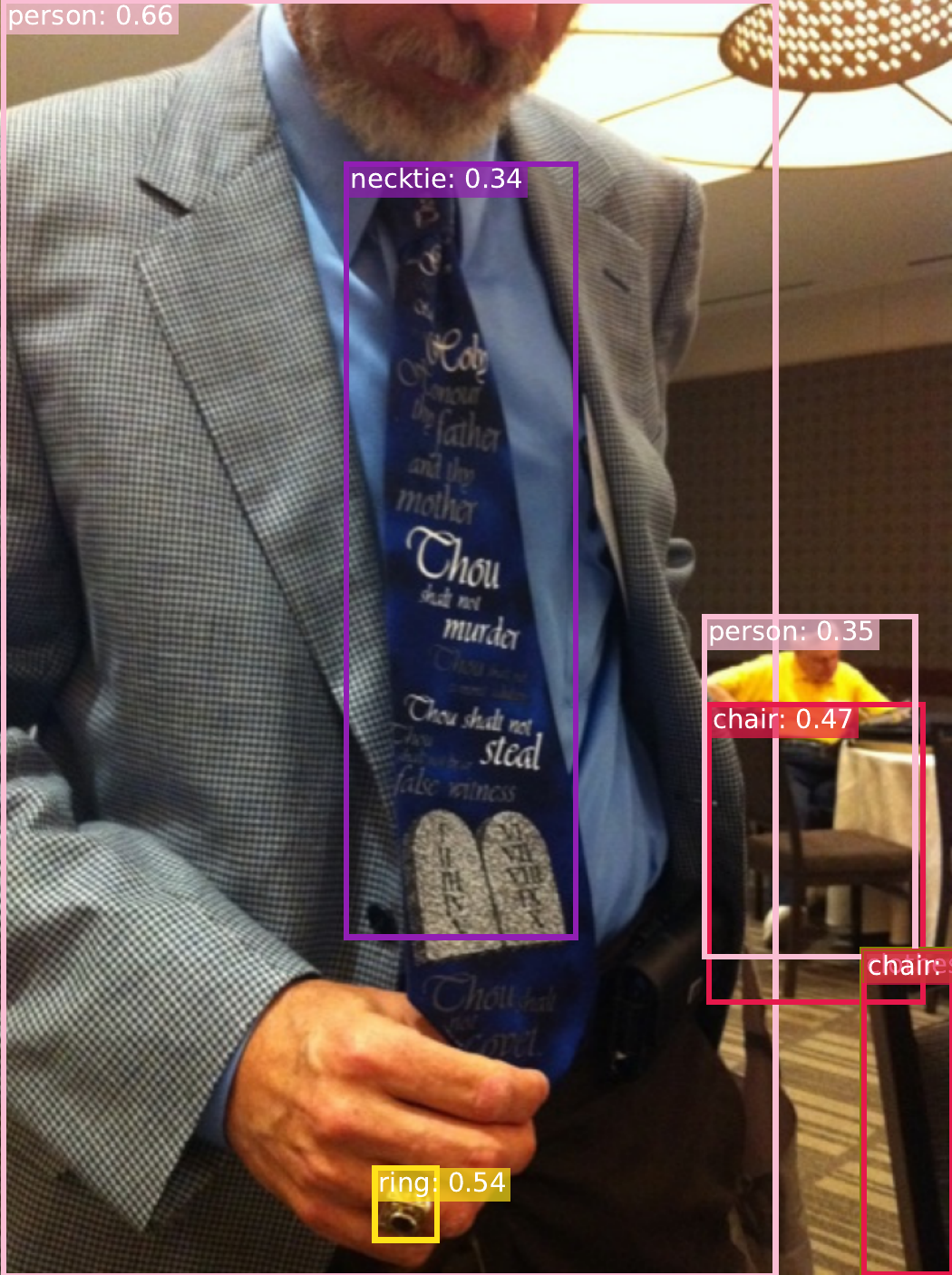} \\
    \end{tabular}
    \caption{Qualitative comparison of LVIS zero-shot detection on YOLO-World-L (4-4-8, Ch-T-H)}
    \label{fig:supp-lvis-vis}
\end{figure}

\fi
\end{document}